\documentclass[twoside]{article}
\usepackage{arxiv}

\usepackage[utf8]{inputenc}
\usepackage{natbib}
\usepackage{bm}
\usepackage{amsmath}
\usepackage{amsfonts}
\usepackage{amssymb}
\usepackage{graphicx}
\usepackage{soul}
\usepackage{subfig}
\usepackage{comment}
\usepackage{multirow}
\usepackage{cases} 
\usepackage{float}

\usepackage{algorithm, algorithmic}
\usepackage{enumitem}
\usepackage{hyperref}

\usepackage[dvipsnames]{xcolor}         
\usepackage{array}
\usepackage{booktabs}
\usepackage{pifont}

\newcommand{\vd}{{\bm{d}}}
\renewcommand{\vd}{{\bm{d}}}
\newcommand{\vx}{{\bm{x}}}
\newcommand{\vX}{{\bm{X}}}

\newcommand{\vtheta}{{\bm{\theta}}}

\renewcommand{\L}{{\mathcal{L}}}
\newcommand{\N}{\mathcal{N}}

\newcommand{\data}{{\mathcal{D}}}

\newcommand{\mtrue}{m_\text{true}}
\newcommand{\thetatrue}{\vtheta_\text{true}}
\newcommand{\mest}{m_\text{est}}
\newcommand{\thetaest}{\vtheta_\text{est}}

\newcommand{\cmark}{\color{OliveGreen} \ding{51}}%
\newcommand{\xmark}{\color{BrickRed} \ding{55}}%

\title{Online simulator-based experimental design for cognitive model selection}

\author{Alexander Aushev \\
  Department of Computer Science\\
  Aalto University, Finland \\
  \texttt{alexander.aushev@aalto.fi} \\
  \And
  Aini Putkonen \\
  Department of Communications and Networking\\
  Aalto University, Finland \\
  \texttt{aini.putkonen@aalto.fi}
  \And
  Gregoire Clarte \\
  Department of Computer Science \\ 
  University of Helsinki \& FCAI, Finland \\
  \texttt{gregoire.clarte@helsinki.fi}
  \And 
  Suyog Chandramouli \\
Department of Communications and Networking\\
  Aalto University, Finland \\
\texttt{suyog.chandramouli@aalto.fi}
  \And
  Luigi Acerbi \\ 
  Department of Computer Science \\ 
  University of Helsinki \& FCAI, Finland \\
  \texttt{luigi.acerbi@helsinki.fi}
  \And 
  Samuel Kaski \\
  Department of Computer Science \\ 
  Aalto University, Finland \\ 
  Department of Computer Science \\ 
  University of Manchester, UK \\
  \texttt{samuel.kaski@aalto.fi} \\
  \And 
  Andrew Howes \\
  School of Computer Science \\
  University of Birmingham, UK \\
  \texttt{a.howes@bham.ac.uk } \\
}

\begin{document}

\maketitle

\begin{abstract}
    The problem of model selection with a limited number of experimental  trials has received considerable attention in cognitive science, where the role of experiments is  to discriminate between theories expressed as computational models. Research on this subject has mostly been restricted to optimal experiment design with analytically tractable models. However, cognitive models of increasing complexity, with intractable likelihoods, are becoming more commonplace. In this paper, we propose BOSMOS: an approach to  experimental design  that can select between computational models without tractable likelihoods. It does so in a data-efficient manner, by sequentially and adaptively generating informative experiments. In contrast to previous approaches, we introduce a novel simulator-based utility objective for design selection, and a new approximation of the model likelihood for model selection. In simulated experiments, we demonstrate that the proposed BOSMOS technique can accurately select models in up to $2$ orders of magnitude less time than existing LFI alternatives for three cognitive science tasks: memory retention, sequential signal detection and risky choice.
\end{abstract}

\section{Introduction}

The problem of selecting between competing models of cognition is critical to progress in cognitive science. The goal of model selection is to choose the model that most closely represents the cognitive process which generated the observed behavioural data. Typically, model selection involves maximizing the fit of each model's parameters to data and balancing the quality of the model-fit and its complexity. It is crucial that any model selection method used is robust and sample-efficient, and that it correctly measures how well each model approximates the data-generating cognitive process. 

It is also crucial that any model selection process is provided with high quality data from well-designed experiments, and that these data are sufficiently informative to support efficient selection. Research on \emph{optimal experimental design} (OED) addresses this problem by focusing on how to design experiments that support parameter estimation of single models and, in some cases, maximize information for model selection \citep{cavagnaro2010adaptive, moon2022speeding, blau2022optimizing}. 

However, one outstanding difficulty for model selection is that many models do not have tractable likelihoods. The model likelihoods represent the probability of  observed data being produced by model parameters and  simplify tractable inference \citep{van2020unbiased}. In their absence, likelihood-free inference (LFI) methods can be used, which rely on forward simulations (or samples from the model) to replace the likelihood. Another difficulty is that existing methods for OED are slow -- very slow -- which makes them impractical for many applications. In this paper, we address these problems by investigating a new algorithm that automatically designs experiments for likelihood-free models much more quickly than previous approaches. The new algorithm is called \emph{Bayesian optimization for simulator-based model selection} (BOSMOS).

In BOSMOS, model selection is  conducted in a Bayesian framework. In this setting, inference  is carried out using \emph{marginal likelihood}, which incorporate, by definition, a penalty for model complexity, \emph{i.e.}, Occam's Razor. Additionally, the Bayesian framework allows getting Bayesian posteriors over all possible values, rather than point estimates; this is crucial for quantifying uncertainty, for instance, when multiple models can explain the data similarly well (non-identifiability or poor identifiability; \citealp{anderson1978arguments,acerbi2014framework}), or when some of the models are misspecified (\emph{e.g.} the behaviour cannot be reproduced by the model due to non-independence of the experimental trials; \citealp{lee2019robust}). 
These problems are compounded in computational cognitive modeling where non-identifiability also arises due to human strategic flexibility \citep{howes2009rational, madsen2019analytic, kangasraasio2019parameter, oulasvirta2022computational}. For these reasons, there is an  interest in Bayesian approaches in computational cognitive science \citep{madsen2018large}.

As we have said, a key problem for model selection is selection of the \emph{design variables} that define an experiment. When resources are limited,  experimental designs can be carefully selected to yield as much information about the models as possible. Adaptive Design Optimization (ADO) \citep{cavagnaro2010adaptive, cavagnaro2013discriminating} is one influential approach to selecting experimental designs. ADO proposes designs by maximizing the so-called utility objective, which measures the amount of information about the candidate models and their quality. Unfortunately, common utility objectives, such as mutual information \citep{shannon1948mathematical, cavagnaro2010adaptive} or expected entropy \citep{yang2005measure}, cannot be applied when computational models lack a tractable likelihood. In such cases, research suggests adopting LFI methods, in which the computational model generates synthetic observations for inference \citep{gutmann2016bayesian, sisson2018handbook, papamakarios2019sequential}. This broad family of methods is also known as approximate Bayesian computation (ABC) \citep{beaumont2002approximate, kangasraasio2019parameter} and simulator- or simulation-based inference \citep{cranmer2020frontier}. To date, LFI methods for ADO have focused on parameter inference for a single model rather than model selection.

Model selection with limited design iterations requires a choice of design variables that optimize model discrimination, as well as improving parameter estimation. The complexity of this task is compounded in the context of LFI, where expensive samples from the model are required. We aim at reducing the number of model simulations. 
For this reason, in our approach, called BOSMOS, we use Bayesian Optimization (BO) \citep{frazier2018bayesian, greenhill2020bayesian} for both design selection and model selection. The advantage of BO is that it is highly sample-efficient and therefore has a direct impact on reducing the need for model simulation. BOSMOS combines the ADO approach with LFI techniques in a novel way, resulting in a faster method to carry out optimal design of experiments to discriminate between computational cognitive models, with a minimal number of trials.

The main contributions of the paper are as follows:
\begin{itemize}
    \item A novel approach to simulator-based model selection that casts LFI for multiple models under the Bayesian framework through the approximation of the model likelihood. As a result, the approach provides a full joint Bayesian posterior for models and their parameters given collected experimental data.
    
    \item A novel simulator-based utility objective for choosing experimental designs that maximizes the behavioural variation in current beliefs about model configurations. Along with the sample-efficient LFI procedure, it reduces the time cost from one hour, for competitor methods, to less than a minute in the majority of case studies, bringing the method closer to enabling real-time cognitive model testing with human subjects.
    
    \item Close integration of the two above contributions yields the first online, sample-efficient, simulation-based, and fully Bayesian experimental design approach to model selection. 
    
    \item The new approach was tested on three well-known paradigms in psychology -- memory retention, sequential signal detection and risky choice -- and, despite not requiring likelihoods, reaches similar accuracy to the existing methods which do require them.
\end{itemize}

\section{Background}
\label{sec:background}

In this article, we are concerned with situations where the purpose of experiments is to gather data that can discriminate between models. The traditional approach in such a context begins with the collection of large amounts of data from a large number of participants on a design that is fixed based on intuition; this is followed by evaluation of the model fits using a desired model selection criteria such as as AIC, BIC, cross-validation, etc. This is an inefficient approach -- the informativeness of the collected data for choosing models is unknown in advance, and collecting large amounts of data may often prove expensive in terms of time and monetary resources (for instance, cases that involve expensive equipment, such as fMRI, or in clinical settings). These issues have been addressed by modern optimal experimental design methods which we consider in this section and summarize in Table \ref{tab:relatedworks}.

\paragraph{Optimal experimental design.} Optimal experiment design (OED) is a classic problem in statistics \citep{lindley1956measure, kiefer1959optimum}, which saw a resurgence in the last decade due to improvements in computational methods and availability of computational resources. Specifically, adaptive design optimization (ADO)  \cite{cavagnaro2010adaptive, cavagnaro2013discriminating} was proposed for cognitive science models, which has been successfully applied in different experimental settings including memory and decision-making. In ADO, the designs are selected according to a global utility objective, which is an average value of the local utility over all possible data (behavioural responses) and model parameters, weighted by the likelihood and priors \citep{myung2013tutorial}. More general approaches, such as \citet{kim2014hierarchical}, improve upon ADO by combining it with hierarchical modelling, which allow them to form richer priors over the model parameters. While useful, the main drawback of these methods is that they work only with tractable (or analytical) parametric models, that is models whose likelihood is explicitly available and whose evaluation is feasible. 

\paragraph{Model selection for simulator-based models.} In the LFI setting, a critical feature of many cognitive models is that they lack a closed-form solution, but allow forward simulations for a given set of model parameters. A few approaches have made advances in tackling the problem of intractability of these models. For instance, \citet{kleinegesse2020bayesian} and \citet{valentin2021bayesian} proposed a method which combines Bayesian optimal experimental design (BOED) and approximate inference of simulator-based models. The Mutual Information Neural Estimation for Bayesian Experimental Design (MINEBED) method performs BOED by maximizing a lower bound on the expected information gain for a particular experimental design, which is estimated by training a neural network on synthetic data generated by the computational model. By estimating mutual information, the trained neural network no longer needs to model the likelihood directly for selecting designs and doing the Bayesian update. Similarly, Mixed Neural Likelihood Estimation (MNLE) by \citet{boelts2022flexible} trains neural density estimators on model simulations to emulate the simulator. \citet{pudlo2016reliable} proposed an LFI approach to model selection, which uses random forests to approximate the marginal likelihood of the models. Despite these advances, these methods have not been designed for model selection in an adaptive experimental design setting. Table \ref{tab:relatedworks} summarizes the main differences between modern approaches and the method proposed in this paper.

Cognitive models increasingly operate in an agent-based paradigm \citep{madsen2019analytic}, where the model is a reinforcement learning (RL) policy \citep{kaelbling1996reinforcement, sutton2018reinforcement}. The main problem with these agent-based models is that they need retraining if any of their parameters are altered, which introduces a prohibitive computational overhead when doing model selection. Recently, \citet{moon2022speeding} proposed a generalized model parameterized by cognitive parameters, which can quickly adapt to multiple behaviours, theoretically bypassing the need for model selection altogether and replacing it with parameter inference. Although the cost of evaluating these models is low in general, they lack the interpretability necessary for cognitive theory development. Therefore, training a parameterized policy within a single RL model family may be preferable: this would still require model selection but would avoid the need for retraining when parameters change (see Section \ref{sec:signal-detecton} for a concrete example).

\paragraph{Amortized approaches to OED.} Recently proposed amortized approaches to OED \citep{blau2022optimizing} -- i.e., flexible machine learning models trained upfront on a large set of problems, with the goal of making fast design selection at runtime -- allow more efficient selection of experimental designs by introducing an RL policy that generates design proposals. This policy provides a better exploration of the design space, does not require access to a differentiable probabilistic model and can handle both continuous and discrete design spaces, unlike previous amortized approaches \citep{foster2021deep, ivanova2021implicit}. These amortized methods are yet to be applied to model selection.

Even though OED is a classical problem in statistics, its application has mostly been relegated to discriminating between simple tractable models. Modern methods such as likelihood-free inference and amortized inference can however make it more feasible to develop OED methods that can work with complex simulator models. In the next sections, we elaborate on our LFI-based method BOSMOS, and demonstrate its working using three classical cognitive science tasks: memory retention, sequential signal detection and risky choice.

\begin{table}[]
    \centering
    \begin{tabular}{m{0.15\textwidth}ccccc}
        \hline
        \textbf{Reference} & \textbf{Method} & \textbf{LFI} & \textbf{Model sel.} & \textbf{Par. inf.} & \textbf{Amortized} \\ \hline
        
        \citet{cavagnaro2010adaptive} & ADO & \xmark & \cmark & \cmark & \xmark \\
        
        \citet{kleinegesse2020bayesian} & MINEBED & \cmark &  \xmark & \cmark &  \cmark \\
        
        
        \citet{blau2022optimizing} & RL-BOED & \cmark  & \xmark  & \cmark  & \cmark \\
        
        \citet{moon2022speeding} & BOLFI & \cmark  & \xmark  & \cmark  & \xmark \\
        
        \citet{pudlo2016reliable} & RF-ABC & \cmark & \cmark & \xmark & \cmark \\
        
        This work  & BOSMOS & \cmark & \cmark & \cmark & \xmark \\
    \end{tabular}
    \caption{Comparison of experimental design approaches to parameter inference (Par. inf.) and model selection (Model sel.) with the references to the selected representative works. Here, we emphasize LFI methods, as they do not need tractable model likelihoods, and amortized methods since they are the fastest to propose designs. The amortized approaches, however, need to be retrained when the population distributions (i.e. priors over models or parameters) change, as in the setting such as ours where beliefs are updated sequentially as new data are collected.}
    \label{tab:relatedworks} 
\end{table}

\section{Methods}
\label{sec:methods}
\begin{figure}
    \centering
    \includegraphics[width=\textwidth]{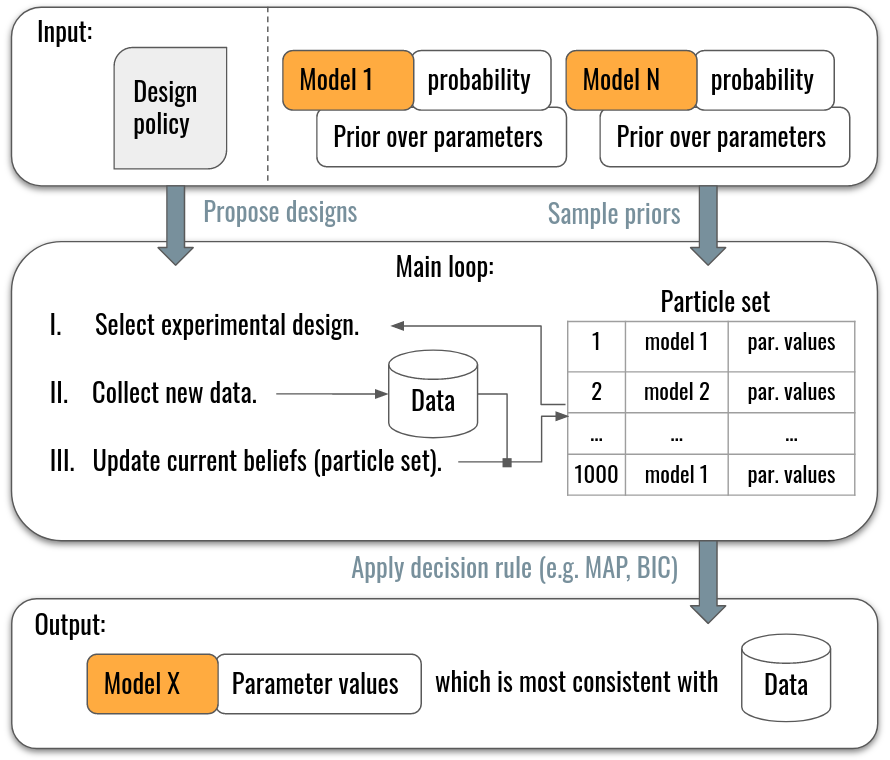}
    \caption{Components of the model selection approach. Main loop continues until the experimental design budget is depleted. Input panel: the experimenter defines a design policy (e.g. random choice of designs), as well as the models and their parameter priors. Middle panel: (i) the next experimental design is selected based on the design policy and current beliefs about models and their parameters (initially sampled from model and parameter priors); (ii) the experiment is carried out using the chosen design, and the observed response-design pair is stored; (iii) current beliefs are updated (e.g. resampled) based on experimental evidence acquired thus far. Output panel: the model and parameters that are most consistent with the collected data are selected by applying one of the well-established decision rules to the final beliefs about models and their parameters.}
    \label{fig:model-selection-overiew}
\end{figure}

Our method carries out optimal experiment design for model selection and parameter estimation involving three main stages as shown in Figure \ref{fig:model-selection-overiew}: selecting the experimental design $\vd$, collecting new data $\vx$ at the design $\vd$ chosen from a design space, and, finally, updating current beliefs about the models and their parameters. The process continues until the allocated budget for design iterations $T$ is exhausted, and the preferred cognitive model $\mest \in \mathcal{M}$, which explains the subject behaviour the best, and its parameters $\thetaest \in \Theta_\text{est}$ are extracted. While the method is rooted in Bayesian inference and thus builds a full joint posterior over models and parameters, we also consider that ultimately the experimenter may want to report the single `best' model and parameter setting, and we use this decision-making objective to guide the choices of our algorithm. The definition of what `best' here means depends on a cost function chosen by the user \citep{robert2007bayesian}. In this paper, for the sake of simplicity, we choose the most common Bayesian estimator, the \emph{maximum a posteriori} (MAP), of the \emph{full posterior} computed by the method: 
\begin{align}
    \mest &= \text{arg max}_m p(m \mid \data_{1:t}), \label{eq:mapm}\\
    \thetaest &= \text{arg max}_{\vtheta_m} p(\vtheta_m \mid m, \data_{1:t}), \label{eq:maptheta}
\end{align}
where $m \in \mathcal{M}$, $\vtheta_m \in \Theta_m$ and $\data_{1:t} = ((\vd_1, \vx_1), ... (\vd_t, \vx_t))$ is a sequence of experimental designs $\vd$ (e.g. shown stimulus) and the corresponding behavioural data $\vx$ (e.g. the response of the subject to the stimuli) pairs.

In our usage context, it is important to make a few reasonable assumptions. First, we assume that the prior over the models $p(m)$ and their parameters $p(\vtheta_m \mid m)$, as well as the domain of the design space, have been specified using sufficient prior knowledge; they may be given by expert psychologists or previous empirical work. This guarantees that the space of the problem is well-defined. Notice that this also implies that the set of candidate models $\mathcal{M} = \left(m_1, \ldots, m_k\right)$ is known, and each model is defined, for any design, by its own parameters. Second, we assume that the computational models that we consider may not necessarily have a closed-form solution, in case their likelihoods $p(\vx \mid \vd, \vtheta_m, m)$ are intractable, 
but it is possible to sample from the forward model $m$, given parameter setting $\vtheta_m$, and design $\vd$. In other words, we operate in a simulator-based inference setting. Please note that this likelihood depends only on the current design and parameters, as assumed in our setting. The third assumption is that each subject's dataset is analyzed separately: we consider single subjects with fixed parameters undergoing the whole set of experiments, as opposed to the statistical setting where information about one dataset may impact the whole population such as, for instance, in hierarchical modelling or pooled models.

As evidenced by Equations \eqref{eq:mapm} and \eqref{eq:maptheta}, the sequential choice of the designs at any point depends on the current posterior over the models  and parameters $p(\vtheta_m, m \mid \data_{1:t}) = p(\vtheta_m \mid \data_{1:t}, m) \cdot p(m \mid \data_{1:t})$, which needs to be approximated and updated at each iteration step of the main loop in Figure \ref{fig:model-selection-overiew}. This problem can be formulated through sequential importance sampling methods, such as Sequential Monte Carlo (SMC; \citealp{del2006sequential}). Thus, the resulting parameter posteriors can be approximated, up to resampling, in the form of equally weighted particle sets:
$q_t(\vtheta_m,m \mid  \data_{1:t}) = \sum_{i=1}^{N_1} N_1^{-1} \delta_{\vtheta^{(i)}_m,m^i}$, with $\vtheta^{(i)}_m,m^{(i)}$ the parameters and models associated with the particle $i$, as an approximation of $p(\vtheta_m \mid m, \data_{1:t})$. These particle sets are later sampled to select designs and update parameter posteriors. In the following sections, we take a closer look at the design selection and belief update stages.

\subsection{Selecting experimental designs}
\label{sec:design}

Traditionally, in the experimental design literature, the designs are selected at each iteration $t$ by maximizing the reduction of the expected entropy $H(\cdot)$ of the posterior $p(m, \vtheta_m \mid \data_{1:t})$. By definition of conditional probability we have:
 \begin{align}
    \vd_t &= \text{argmin}_{\vd_t} \mathbb{E}_{\vx_t \mid \data_{1:t-1}} \big[H(\vtheta_m, m \mid \data_{1:t-1} \cup (\vd_t, \vx_t)) \big] \label{eq:designloss} \\
    &= \text{argmin}_{\vd_t} \mathbb{E}_{\vx_t \mid \data_{1:t-1}} \left[ \mathbb{E}_{p(\vtheta_m, m \mid \data_{1:t})} [-\log p(\vtheta_m, m \mid \data_{1:t} \cup (\vd_t, \vx_t))] \right] \nonumber\\
    &=\mathrm{argmin}_{\vd_t}\mathbb{E}_{\vx_t \mid \data_{1:t-1}}\mathbb{E}_{p(\vtheta_m, m \mid \data_{1:t-1})} \left[-\log(p(\vx_t \mid \vd_t, \vtheta_m, m)) \right] \nonumber \\ 
    &\hspace{1cm} + \mathbb{E}_{\vx_t\mid\data_{1:t-1}} \log p(\vx_t \mid \vd_t, \data_{1:t-1}) \label{eq:designloss-2},  
\end{align}
where $\vx_t$ is the response predicted by the model. The first equality comes from the definition of entropy and the second from Bayes rule, where we removed the prior, as this term is a constant term in $\vd_t$. Here, lower entropy corresponds to a narrower, more concentrated, posterior -- with maximal information about models and parameters.

Since neither $p(\vx_t \mid \vd_t, \vtheta_m, m)$ nor, by extension, Equation \eqref{eq:designloss-2} are tractable in our setting, we propose a simulator-based utility objective
\begin{align}
    \label{eq:heurloss_prat}
    \vd_t &= \text{arg min}_{\vd_t} \mathbb{E}_{q_t(\vtheta_m, m \mid \data_{1:t-1})} [\hat{H}(\vx_t' \mid \vd_t, \vtheta_m,m)]  - \hat{H}(\vx_t \mid D_{1:t-1},\vd_t),
\end{align}
where $q_t$ is a particle approximation of the posterior at time $t$, and $\hat{H}$ is a  kernel-based Monte Carlo approximation of the entropy $H$. 

The intuition behind this utility objective is that we choose such designs $\vd_t$ that would maximize identifiability (minimize the entropy) between $N$ responses $\vx'$ simulated from different computational models $p(\cdot \mid \vd_t, \vtheta_m, m)$. The models $m$ as well as their parameters $\vtheta_m$ are sampled from the current beliefs $q_t(\vtheta_m, m \mid \data_{1:t-1})$. The full asymptotic validity of the Monte Carlo approximation of the decision rule in Equation \eqref{eq:heurloss_prat} can be found in Appendix A.

The utility objective in \eqref{eq:heurloss_prat} allows us to use Bayesian Optimization (BO) to find the design $\vd_t$ and then run the experiment with the selected design. In the next section, we discuss how to update beliefs about the models $m$ and their parameters $\vtheta_m$ based on the data collected from the experiment.

\subsection{Likelihood-free posterior updates}
\label{sec:lfi}

The response $\vx_t$ from the experiment with the design $\vd_t$ is used to update approximations of the posterior $q_t(m \mid \data_t)$ and $q_t(\vtheta_m \mid m, \data_t)$, obtained via marginalization and conditioning, respectively, from $q_t(\vtheta_m, m \mid \data_t)$. We use LFI with synthetic responses $\vx_{\vtheta_m}$ simulated by the behavioural model $m$ to perform the approximate Bayesian update.

\paragraph{Parameter estimation conditioned on the model.} We start with parameter estimation for each of the candidate models using Bayesian Optimization for Likelihood-Free Inference (BOLFI; \citealp{gutmann2016bayesian}). In BOLFI, a Gaussian process (GP) \citep{rasmussen2003gaussian} surrogate for the discrepancy function between the observed and simulated data, $\rho(\vx_{\vtheta_m}, \vx_t)$ (e.g., Euclidean distance), serves as a base to an unnormalized approximation of the intractable likelihood $p(\vx_t \mid \vd_t, \vtheta_m, m)$. Thus, the posterior can be approximated through the following approximation of the likelihood function $\mathcal{L}_{\epsilon_m}(\cdot)$ and the prior over model parameters $p(\vtheta_m)$:
\begin{align}
    p( \vtheta_m \mid \vx_t) &\propto \mathcal{L}_{\epsilon_m}(\vx_t \mid \vtheta_m) \cdot p(\vtheta_m),
    \label{eq:post} \\
    \mathcal{L}_{\epsilon_m}( \vx_t \mid\vtheta_m) &\approx \mathbb{E}_{\vx_{\vtheta_m}} [\kappa_{\epsilon_m} ( \rho_m(\vx_{\vtheta_m}, \vx_t) )]. \label{eq:liklhd}
\end{align}
Here, following Section 6.3 of \citep{gutmann2016bayesian}, we choose $\kappa_{\epsilon_m}(\cdot) = \mathbf{1}_{[0,\epsilon_m]}(\cdot)$, where the bandwidth $\epsilon_m$ takes the role of a acceptance/rejection threshold. Using a Gaussian likelihood for the GP, this leads to: $\mathbb{E}_{\vx_{\vtheta_m}}[\kappa_{\epsilon_m}(\rho(\vx_{\vtheta_m}, \vx_t))] = \Phi( (\epsilon_m - \mu(\vtheta_m)) / \sqrt{\nu(\vtheta_m) + \sigma^2})$, where $\Phi(\cdot)$ denotes the standard Gaussian cumulative distribution function (cdf). Note that $\mu(\vtheta_m)$ and $\nu(\vtheta_m) + \sigma^2$ are the posterior predictive mean and variance of the GP surrogate at $\vtheta_m$.

\paragraph{Model estimation.} A principled way of performing model selection is via the marginal likelihood, that is $p(\vx_t \mid m) = \int p(\vx_t \mid \vtheta_m,m) \cdot p(\vtheta_m \mid m) \mathrm{d}\vtheta_m$, which is proportional to the posterior over models assuming an equal prior for each model. Unfortunately, a direct computation of the marginal likelihood is not possible with Equation \eqref{eq:liklhd}, since it only allows us to compute a likelihood approximation up to a scaling factor that implicitly depends on $\epsilon$. For instance, when calculating a Bayes factor (ratio of marginal likelihoods) for models $m_1$ and $m_2$
\begin{align}
    \label{eq:bfactor}
    \frac{p(\vx_t \mid m_1)}{p(\vx_t \mid m_2)} = \frac{\mathbb{E}_{\vtheta_{m1}} [p(\vx_t \mid \vtheta_{m1}, m_1)]}{\mathbb{E}_{\vtheta_{m2}}[p(\vx_t \mid \vtheta_{m2}, m_2)]} \neq \frac{\mathbb{E}_{\vtheta_{m1}} [\L_{\epsilon_{m1}}(\vx_t \mid \vtheta_{m1})]}{\mathbb{E}_{\vtheta_{m2}}[\L_{\epsilon_{m2}}(\vx_t \mid \vtheta_{m2})]},
\end{align}
their respective $\epsilon_{m1}$ and $\epsilon_{m2}$, chosen independently, may potentially bias the marginal likelihood ratio in favour of one of the models, rendering it unsuitable for model selection. Choosing the same $\epsilon$ for each model is not possible either, as it would lead to numerical instability due to the shape of the kernel.

To approximate the marginal likelihood $p(\vx_t \mid m)$, we adopt a similar approach as in Equation \eqref{eq:liklhd}, by reframing the marginal likelihood computation as a distinct LFI problem.
In ABC for parameter estimation, we would generate pseudo-observations from the prior predictive distribution of each model, and compare the discrepancy with the true observations on a scale common to all models. This comparison involves a kernel that maps the discrepancy into a likelihood approximation. For example, in rejection ABC \citep{tavare1997inferring, marin2012approximate} this kernel is uniform. In our case, we will generate samples from the joint prior predictive distribution on both models and parameters, and we use a Gaussian kernel $\kappa_\eta(\cdot) = \mathcal{N}(\cdot \mid 0, \eta^2)$, chosen to satisfy all of the requirements from \citet{gutmann2016bayesian}; in particular, this kernel is non-negative, non-concave and has a maximum at $0$. The parameter $\eta > 0$ serves as the kernel bandwidth, similarly to $\epsilon_m$ in Equation~\eqref{eq:liklhd}. The value of $\kappa_\eta(\cdot)$ monotonically increases as the model $m$ produces smaller discrepancy values. This kernel leads to the following approximation of the marginal likelihood:
\begin{align}
    \label{eq:kernel-approx}
    \L(\vx_t \mid m, \data_{t-1}) &\propto  \mathbb{E}_{ \vx_{\vtheta} \sim p(\cdot \mid \vtheta_m, m) \cdot q(\vtheta_m \mid m, \data_{t-1}) } \kappa_\eta(\hat{\rho}(\vx_{\vtheta}, \vx_t)),
\end{align}
where $\kappa_\eta(\cdot) = \mathcal{N}(\cdot \mid 0, \eta^2)$, and $\hat{\rho}$ is the GP surrogate for the discrepancy. Eq. \ref{eq:kernel-approx} is a direct equivalent of Eq. \ref{eq:liklhd}, but here we integrate (marginalize) over both $\theta$ and $x_\theta$. Here we used the Gaussian kernel, instead of the uniform kernel used in Eq. \ref{eq:liklhd}, as it produced better results for model selection in preliminary numerical experiments. Note that in Eq. \ref{eq:kernel-approx} we have two approximations, the first one from $\kappa_\eta$, stating that the likelihood is approximated from the discrepancy, and the second from the use of a GP surrogate for the discrepancy.

The choice of $\eta$ is a complex problem, and in this paper we propose the simple solution of setting $\eta$ as the minimum value of $\mathbb{E}_{ \vx_{\vtheta} \sim p(\cdot \mid \vtheta_m, m) \cdot q(\vtheta_m \mid m, \data_{t-1}) }\hat{\rho}(\vx_{\vtheta}, \vx_t)$ across all models $m \in \mathcal{M}$. This value has the advantage of giving non extreme values to the estimations of the marginal likelihood, which should in principle avoid over confidence.

\paragraph{Posterior update.} The resulting marginal likelihood approximation in Equation~\eqref{eq:kernel-approx} can then be used in posterior updates for new design trials as follows:
\begin{align}
    q(m \mid \data_t) &\propto \L(\vx_t \mid m, \data_{t-1}) \cdot q(m\mid \data_{t-1}) \approx \kappa_\eta(\omega_m) \cdot q(m \mid \data_{t-1}), \\
    q(\vtheta_m \mid m,\data_t) &\propto \L_{\epsilon_m}(\vx_t \mid \vtheta_m, m) \cdot q(\vtheta_m \mid \data_{t-1},m).
\end{align}

Which is equivalent to:
\begin{equation}
    \label{eq:update}
    q(\vtheta_m, m \mid \data_t) \propto \L_{\epsilon_m}(\vx_t \mid \vtheta_m, m) \cdot \L(\vx_t \mid m, \data_{t-1}) \cdot q(\vtheta_m, m \mid \data_{t-1}).
\end{equation}

Once we updated the joint posterior of models and parameters, it is straightforward to obtain the model and parameter posterior through marginalization and apply a decision rule (e.g. MAP) to choose the estimate. The entire algorithm for BOSMOS can be found in Appendix B.


\section{Experiments}

\begin{figure}
    \centering
    \includegraphics[width=\textwidth]{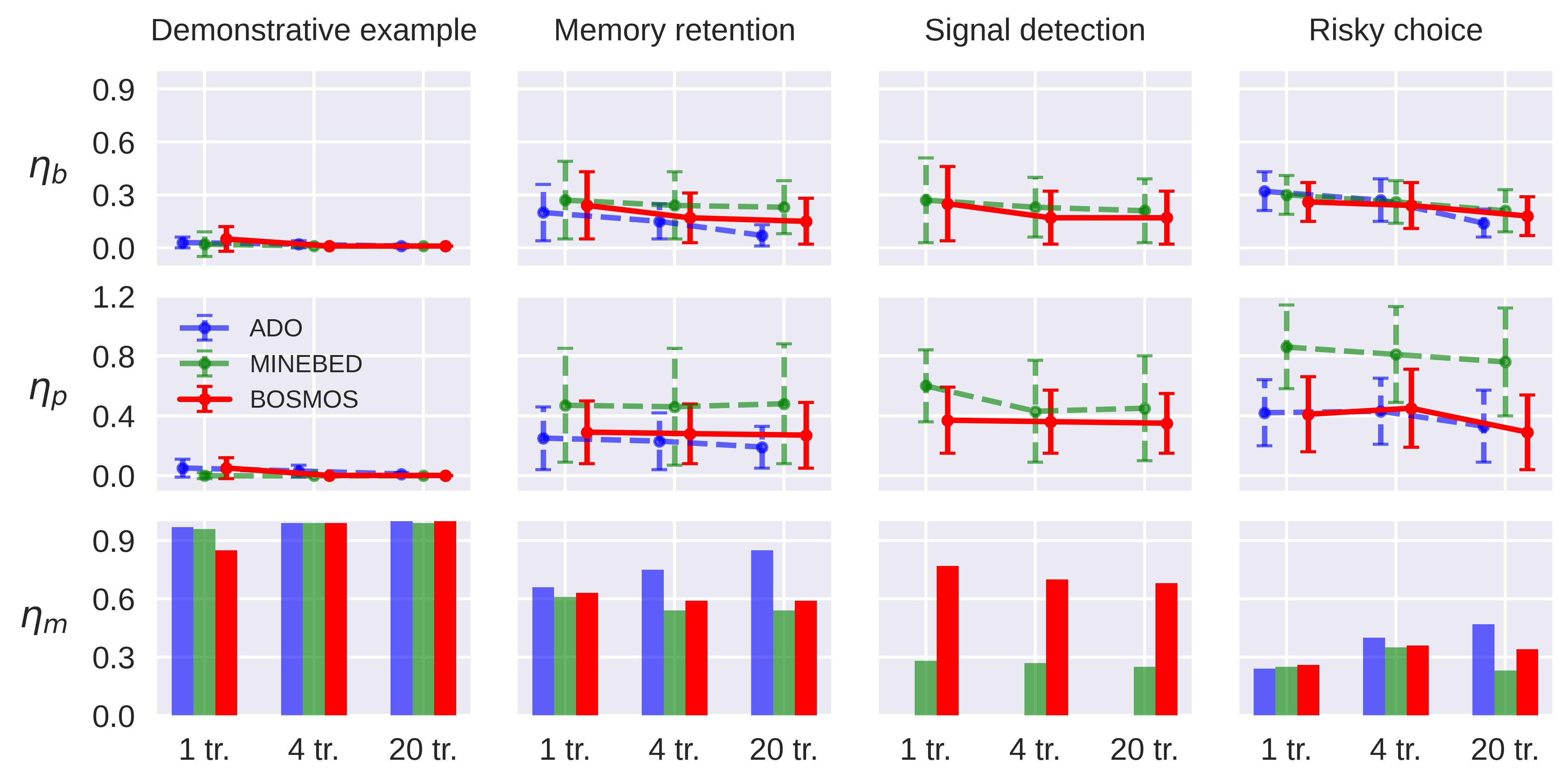}
    \caption{Evaluation of three performance measures (rows) after 1, 4 and 20 design trials for BOSMOS (solid red) and two alternative best methods, ADO (blue) and MINEBED (green), in four cognitive tasks (columns).  As the number of design trials grows, the methods accumulate more observed data from subjects' behaviour and, hence, should reduce behavioural fitness error $\eta_\text{b}$, parameter estimation error $\eta_\text{p}$, and increase model predictive accuracy $\eta_\text{m}$. Since $\eta_\text{b}$ is the performance metric MINBED and BOSMOS optimize, its convergence is the most prominent. The lack of convergence for the other two metrics in the memory retention and signal detection tasks is likely due to the possibility of the same behavioural data being produced by models and parameters that are different from the ground-truth (\emph{i.e.}, non-identifiability of these models).}
    \label{fig:conv}
\end{figure}

In the experiments, our goal was to evaluate how well the proposed method described in Section \ref{sec:methods} discriminated between different computational models in a series of cognitive tasks: memory retention, signal detection and risky choice.  Specifically, we measured how well the method chooses designs which help the estimated model imitate the behaviour of the target model, discriminate between models, and correctly estimate their ground-truth parameters. In our simulated experimental setup, we created $100$ synthetic participants by sampling the ground-truth model and its parameters (not available in the real world) through priors $p(m)$ and $p(\vtheta_m \mid m)$. Then, we ran the sequential experimental design procedure for a range of methods described in Section \ref{sec:comparisonmethods}, and recorded four main performance metrics shown in Figure \ref{fig:conv} for $20$ design trials (results analysed further later in the section): the behavioural fitness error $\eta_\text{b}$, defined below, the parameter estimation error $\eta_\text{p}$, the accuracy of the model prediction $\eta_\text{m}$ and the empirical time cost of running the methods. Furthermore, we evaluated the methods at different stages of design iterations in Figure \ref{fig:conv} for the convergence analysis. The complete experiments with additional evaluation points can be found in Appendix C.
  
We compute $\eta_\text{b}$, $\eta_\text{p}$ and $\eta_\text{m}$ for a single synthetic participant using the known ground truth model $\mtrue$ and parameters $\thetatrue$. The behavioural fitness error $\eta_\text{b}=\Vert \vX_\text{true} - \vX_\text{est} \Vert^2$
is calculated as the Euclidean distance between the ground-truth model ($\vX_\text{true}$) and synthetic ($\vX_\text{est}$) behavioural datasets, which consist of means $\mu(\cdot)$ of $100$ responses evaluated at the same $100$ random designs $\mathcal{T}$ generated from a proposal distributions $p(\vd)$, defined for each model: 
\begin{align}
    \mathcal{T} &= \{ \vd_i \sim p(\vd)\}_{i=1}^{100}, \\
    \vX_\text{true} &= \{ \mu(\{ \vx_s:  \vx \sim p(\cdot \mid \vd_i, \thetatrue, \mtrue) \}_{s=1}^{100}): \vd_i \in \mathcal{T}\}_{i=1}^{100}, \\
    \vX_\text{est} &= \{ \mu(\{ \vx_s:  \vx \sim p(\cdot \mid\vd_i, \thetaest, \mest) \}_{s=1}^{100}): \vd_i \in \mathcal{T} \}_{i=1}^{100}.
\end{align}
Here, $\mest$ and $\thetaest$ are, respectively, the model and parameter values  estimated via the MAP rule (unless specified otherwise). $\mest$ is also used to calculate the predictive model accuracy $\eta_\text{m}$ as a proportion of correct model predictions for the total number of synthetic-participants, while $\thetaest$ is used to calculate the averaged Euclidean distance $\Vert \thetatrue - \thetaest \Vert^2$ across all synthetic participants, which constitutes the parameter estimation error $\eta_\text{p}$.

\begin{figure}
    \centering
    \includegraphics[width=\textwidth]{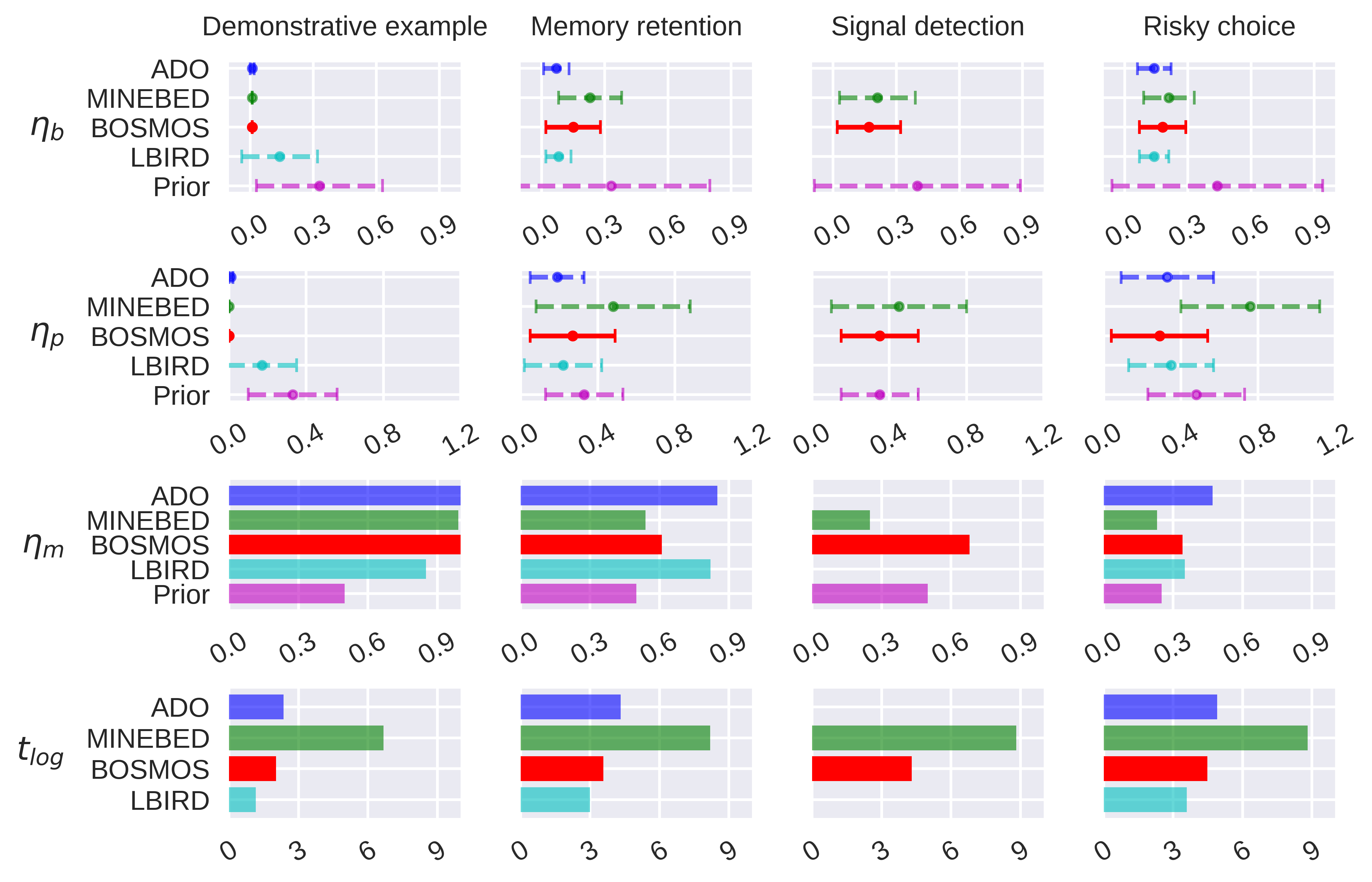}
    \caption{An overview of the performance of the methods, compared with the prior predictive with random design, (rows) after 20 design trials across four different cognitive modelling tasks (columns): demonstrative example, memory retention, signal detection and risky choice. While requiring $10$ times fewer simulations and $60$-$100$ times less time, the proposed BOSMOS method (red) shows consistent improvement over the alternative LFI method, MINEBED (green), in terms of behavioural fitness error $\eta_\text{b}$, parameter estimation error $\eta_\text{p}$, model predictive accuracy $\eta_\text{m}$ and empirical time cost $t_{log}$ (here, for $100$ designs, in minutes on a log scale). The model accuracy bars indicate the proportion of correct prediction of models across $100$ simulated participants. The error bars show the mean (marker) and std. (caps) of the error by the respective methods. }
    \label{fig:res-20}
\end{figure}

\subsection{Comparison methods}
\label{sec:comparisonmethods}

Throughout the experiments, we compare several strategies for experimental design selection and parameter inference, where prior predictive distribution (evaluation of the prior without any collected data) with random design choice from the proposal distribution of each model is used as a baseline (we call this method results Prior in the results). The explanations of these methodologies, as well as the exact setup parameters, are provided below.

\subsubsection{Likelihood-based inference with random design}

“Likelihood-based” inference with random design (LBIRD) applies the ground-truth likelihood, where it is possible, to conduct Bayesian inference and samples the design from the proposal distribution $p(\vd)$ instead of design selection: $\data_t= (x_t,\vd_t), \ x_t \sim \pi(\cdot \mid \vtheta,m,\vd_t), \ \vd_t \sim p(\cdot)$. 
This procedure serves as a baseline by providing unbiased estimates of models and parameters. As other methods in this section, LBIRD uses $5000$ particles (empirical samples) to approximate the joint posterior of models and parameters for each model. The Bayesian updates are conducted through importance-weighted sampling with added Gaussian noise applied to the current belief distribution.

\subsubsection{ADO}

ADO requires a tractable likelihood of the models, and hence is used as an upper bound of performance in cases where the likelihood is available. ADO \citep{cavagnaro2010adaptive} employs BO for the mutual information utility objective: 
\begin{align}
    U(\vd) = \sum_{m=1}^K p(m) \sum_y p(\vx \mid m, \vd) \cdot \log\left(\frac{p(\vx \mid m,\vd)}{\sum_{m=1}^K p(m) p(\vx \mid m, \vd)}\right),
\end{align}
where we used $500$ parameters sampled from the current beliefs to integrate
\begin{align}
    p(\vx \mid m, \vd) = \int p(\vx \mid \vtheta_m, m, \vd) \cdot p(\vtheta_m \mid m) \text{d}\vtheta.
\end{align}
Similarly to other approaches below which also use BO, the BO procedure is initialized with $10$ evaluations of the utility objective with $\vd$ sampled from the design proposal distribution $p(\vd)$, while the next $5$ design locations are determined by the Monte-Carlo-based noisy expected improvement objective. The GP surrogate for the utility uses a constant mean function, a Gaussian likelihood and the Matern kernel with zero mean and unit variance. All these components of the design selection procedure were implemented using the BOTorch package \citep{balandat2020botorch}.

\subsubsection{MINEBED}

MINEBED \citep{kleinegesse2020bayesian} focuses on design selection for parameter inference with a single model. Since our setting requires model selections and by extension working with multiple models, we compensate for that by having a separate MINEBED instance for each of the models and then assigning a single model (sampled from the current beliefs) for design optimization at each trial. The model is assigned by the MAP rule over the current beliefs about models $q(m \mid \data_{1:t})$, and the data from conducting the experiment with the selected design are used to update all MINEBED instances. We used the original implementation of the MINEBED method by \citet{kleinegesse2020bayesian}, which uses a neural surrogate for mutual information consisting of two fully connected layers with 64 neurons. This configuration was optimized using Adam optimizer \citep{kingma2014adam} with initial learning rate of $0.001$, $5000$ simulations per training at each new design trial and $5000$ epochs.

\subsubsection{BOSMOS}

BOSMOS is the method proposed in this paper and  described in Section \ref{sec:methods}. It uses the simulator-based utility objective from Equation \eqref{eq:heurloss_prat} in BO to select the design and BO for LFI, along with the marginal likelihood approximation from Equation \eqref{eq:kernel-approx} to conduct inference. The objective for design selection is calculated with the same $10$ models (a higher number increases belief representation at the cost of more computations) sampled from the current belief over models (i.e. particle set $q_t(m \mid \data_{1:t})$ at each time $t$), where each model is simulated $10$ times to get one evaluation point of the utility ($100$ simulations per point). In total, in each iteration, we spent $1500$ simulations to select the design and additional $100$ simulations to conduct parameter inference. 

As for parameter inference in BOSMOS, BO was initialized with $50$ parameter points randomly sampled from the current beliefs about model parameters (i.e. the particle set $q_t(\vtheta_m \mid m, \data_{1:t})$), the other $50$ points were selected for simulation in batches of $5$ through the Lower Confidence Bound Selection Criteria \citep{srinivas2009gaussian} acquisition function. Once again, a GP is used as a surrogate, with the constant mean function and the radial basis function \citep{seeger2004gaussian} kernel with zero mean and unit variance. Once the simulation budget of $100$ is exhausted, the parameter posterior is extracted through an importance-weight sampling procedure, where the GP surrogate with the tolerance threshold set at a minimum of the GP mean function \citep{gutmann2016bayesian} acts as a base for the simulator parameter likelihood.

\subsection{Demonstrative example}
\label{sec:demonstrative-example}

The demonstrative example serves to highlight the significance of design optimization for model selection with a simple toy scenario. We consider two normal distribution models with either positive (PM) or negative (NM) mean. Responses are produced according to the experimental design $d \in [0.001, 5]$ which determines the quantity of observational noise variance:
\begin{align}
    \mathrm{(PM)} \quad x &\sim \N(\theta_\mu, d^2), \\
    \mathrm{(NM)} \quad x &\sim \N(-\theta_\mu, d^2).
\end{align}
These two models have the same prior over parameters $\theta_\mu \in [0, 5]$ and may be clearly distinguished when the optimal design value is $d=0.001$. We choose a uniform prior over models.

\subsubsection{Results}

As shown in the first set of analyses in Figure \ref{fig:res-20}, selecting informative designs can be crucial. When compared to the LBIRD method, which picked designs at random, all the design optimization approaches performed exceedingly well. This highlights the significance of design selection, as random designs produce uninformative results and impede the inference procedure.

Figure \ref{fig:conv} illustrates the convergence of the key performance measures, demonstrating that the design optimization methods had nearly perfect estimates of ground-truths after only one design trial. This indicates that the PM and NM models are easily separable, provided informative designs. In terms of the model predictive accuracy, MINEBED outperformed BOSMOS after the first trial, however BOSMOS rapidly caught up as trials proceeded. This is most likely because our technique employs fewer simulations per trial but a more efficient LFI surrogate than MINEBED. As a result, our method has the second-best time cost not only for the demonstrative example but also across all cognitive tasks. The only method that was faster is the LBIRD method, which skips the design optimization procedure entirely and avoids lengthy computations related to LFI by accessing the ground-truth likelihood.

\subsection{Memory retention}
\label{sec:memory-retention}

Studies of memory are a fundamental research area in experimental psychology.
Memory can be viewed functionally as a capability to encode, store and remember, and neurologically as a collection of neural connections \citep{amin2013HumanMR}.
Studies of memory retention have a long history in psychological research, in particular in relation to the shape of the retention function \citep{rubin1996memoryretention}.
These studies on functional forms of memory retention seek to quantitatively answer how long a learned skill or material is available \citep{rubin1999precise}, or how quickly it is forgotten.
Distinguishing retention functions may be a challenge \citep{rubin1999precise}, and \citet{cavagnaro2010adaptive} showed that employing an ADO approach can be advantageous.
Specifically, studies of memory retention typically consist of a `study phase' (for memorizing) followed by a `test phase' (for recalling), and the time interval between the two is called a `lag time'. 
Varying the lag time by means of ADO allowed more efficient differentiation of the candidate models \citep{cavagnaro2010adaptive}.
To demonstrate our approach with the classic memory retention task, we consider the case of distinguishing two functional forms, or models, of memory retention, defined as follows.

\paragraph{Power and exponential models of memory retention.}
In the classic memory retention task, the subject recalls a stimulus (e.g. a word) at a time $d \in [0, 100]$, which is modelled by two Bernoulli models $B(1, p)$: the power (POW) and exponential (EXP) models. The samples from these models are the responses to the task $x$, which can be interpreted as `stimulus forgotten' in case $x=0$ and $x=1$ otherwise. We follow the definition of these models by \citet{cavagnaro2010adaptive}, where $p = \theta_a \cdot (d+1)^{-\theta_\text{POW}}$ in POW and $p = \theta_a\cdot e^{-\theta_\text{EXP}\cdot d}$ in EXP, as well as the same priors:
\begin{align}
    \theta_a &\sim \text{Beta}(2, 1), \\
    \theta_\text{POW} &\sim \text{Beta}(1, 4), \\
    \theta_\text{EXP} &\sim \text{Beta}(1, 8).
\end{align}
Similarly to the previous demonstrative example and the rest of the experiments, we use equal prior probabilities for the models.

\subsubsection{Results}

Studies on the memory task show that the performance gap between LFI approaches and methods that use ground-truth likelihood grows as the number of design trials increases (Figure \ref{fig:res-20}). This is expected, since doing LFI introduces an approximation error, which becomes more difficult to decrease when the most uncertainty around the models and their parameters has been already removed by previous trials. Unlike in the demonstrative example, where design selection was critical, the ground-truth likelihood appears to have a larger influence than design selection for this task, as evidenced by the similar performance of the LBIRD and ADO approaches. 

In regard to LFI techniques, BOSMOS outperforms MINEBED in terms of behavioural fitness and parameter estimation, as shown in Figure \ref{fig:conv}, but only marginally better for model selection. Moreover, both approaches seem to converge to the wrong solutions (unlike ADO), as evidenced by their lack of convergence in the parameter estimation and model accuracy plots. Interestingly, both techniques continued improving behavioural fitness, implying that behavioural data of the models can be reproduced by several parameters that are different from the ground-truth, and LFI methods fail to distinguish them. A deeper examination of the parameter posterior can reveal this issue, which can be likely alleviated by adding new features for observations and designs that can assist in capturing the intricacies within the behavioural data.

\subsection{Sequential signal detection}
\label{sec:signal-detecton}

Signal detection theory (SDT) focuses on perceptual uncertainty, presenting a framework for studying decisions under such ambiguity \citep{tanner1954sdt, peterson1954sdt, swets1961decision, wickens2002elementary}.
SDT is an influential developing model stemming from mathematical psychology and psychophysics, providing an analytical framework for assessing optimal decision-making in the presence of ambiguous and noisy signals.
The origins of SDT can be traced to the 1800s, but its modern form emerged in the latter half of the 20th century, with the realization that sensory noise is consciously accessible \citep{wixted2020forgotten}.
Example of a signal detection task could be a doctor making a diagnosis: they have to make a decision based on a (noisy) signal of different symptoms \citep{wickens2002elementary}.
SDT is largely considered a normative approach, assuming that a decision-maker is bounded rational \citep{swets1961decision}.
We will consider a sequential signal detection task and two models, Proximal Policy Optimization (PPO) and Probability Ratio (PR), implemented as follows.

\paragraph{SDT.}
In the signal detection task, the subject needs to correctly discriminate the presence of the signal $o_\text{sign} \in \{ \text{present}, \text{absent} \}$ in a sensory input $o_\text{in} \in \mathbb{R}$. The sensory input is corrupted with sensory noise $\sigma_\text{sens} \in \mathbb{R}$:
\[o_\text{in} = 1_{\text{present}} (o_\text{sign}) \cdot d_\text{str} + \gamma, \quad \gamma \propto \mathcal{N}(0, \sigma_\text{sens}). \]
Due to the noise in the observations, the task may require several consecutive actions to finish. At every time-step, the subject has three actions $a \in \{ \text{present}, \text{absent}, \text{look} \}$ at their disposal: to make a decision that the signal is \textit{present} or \textit{absent}, and to take another \textit{look} at the signal. The role of the experimenter is to adjust the signal strength $d_\text{str} \sim \text{Unif}(0, 4)$ and discrete number of observations $d_\text{obs} \sim \text{Unif}_\text{discr}(2, 10)$ the subject can make such that the experiment will reveal characteristics of human behaviour. In particular, our goal is to identify the \textit{hit} value parameter of the subject, which determines how much reward $r(a, s)$ the subject receives, in case the signal is both present and identified correctly.
Hence, we have that

\[ r(a, s) = r_a(s) + r_\text{step}, \]

\begin{align*}
r_a(s) &= \theta_\text{hit}, &\text{when the signal is present, and the action is \textit{present}.} \\
r_a(s) &= 2, &\text{when the signal is absent, and the action is \textit{absent}.} \\
r_a(s) &= 0, &\text{when the action is \textit{look}.} \\
r_a(s) &= -1, &\text{in other cases.} 
\end{align*}
where $r_\text{step} = -0.05$ is the constant cost of every consecutive action.

\paragraph{PPO.} We implement the SDT task as an RL model due to the sequential nature of the task.
In particular, the \emph{look} action will postpone the signal detection decision to the next observation.
The model assumes that the subject acts according to the current observation $o_\text{in}$ and an internal state $\beta$: $\pi(a \mid o_\text{in}, \beta)$.
The internal state $\beta$ is updated over trials by aggregating observations $o_\text{in}$ using a Kalman Filter, and after each trial, the agent chooses a new action.
As we have briefly discussed in Section \ref{sec:background}, the RL policies need to be retrained when their parameters  change. To address this issue, the policy was parameterized and trained using a wide range of model parameters as policy inputs. The resulting model was implemented using the PPO algorithm \citep{schulman2017ppo}.

\paragraph{PR.} An alternative to the RL model is a PR model.
It also assumes sequential observations: a hypothesis test as to whether the signal is present is performed after every observation, and the sequence of observations is called \emph{evidence} \citep{griffith2021statistics}.
A likelihood for the evidence (sequence of observations) is the product of likelihoods of each observation.
\emph{A likelihood ratio} is used as a decision variable (denoted $f_t$ here). 
Specifically, $f_t$ is evaluated against a threshold, which determines which action $a_t$ to take as follows:
\begin{align}
    a_t &= \text{present}, &\text{ if } f_t \leq \theta_\text{low}, \\
    a_t &= \text{absent}, &\text{ if } f_t \geq \theta_\text{low} + \theta_\text{len}, \\
    a_t &= \text{look}, &\text{ if } \theta_\text{low} \leq f_t \leq \theta_\text{low} + \theta_\text{len}. 
\end{align}
where
\begin{align}
    f_t = \prod_{i=1}^{d_\text{obs}} \frac{\omega_1}{\omega_2}, \quad \omega_1 &\sim \N_\text{CDF}\left(\frac{1}{\theta_\text{hit} - 1} ; d_\text{str}, \theta_\text{sens}\right), \\ 
    \quad \omega_2 &\sim \N_\text{CDF}\left( \frac{1}{\theta_\text{hit} - 1} ; 0,  \theta_\text{sens}\right).
\end{align}
Here, $\N_\text{CDF}(\cdot; \mu, \nu)$ is the Gaussian cumulative distribution function (CDF) with the mean $\mu$ and standard deviation $\nu$. For more information about the PR model, we refer the reader to \citet{griffith2021statistics}.

For both models, we used the following priors for their parameters and design values:
\begin{align}
    \theta_\text{sens} &\sim \text{Unif}(0.1, 1), & \theta_\text{hit} &\sim \text{Unif}(1, 7), \\
    \theta_\text{low} &\sim \text{Unif}(0, 5), & \theta_\text{len} &\sim \text{Unif}(0, 5).
\end{align}

\subsubsection{Results}

BOSMOS and MINEBED are the only methodologies capable of performing model selection in sequential signal detection models, as specified in Section \ref{sec:signal-detecton}, due to the intractability of their likelihoods. The experimental conditions are therefore very close to those in which these LFI approaches are usually applied, with the exception that we now know the ground-truth of synthetic participants for performance assessments.

BOSMOS showed a faster convergence of the estimates than MINEBED requiring only 4 design trails to reduce the majority of the uncertainty associated with model prediction accuracy and behaviour fitness error, as demonstrated in Figure \ref{fig:conv}. In contrast, it took $20$ design trials for MINEBED to converge, and extending it beyond $20$ trials provided very little benefit. Similarly as in the memory retention task from Section \ref{sec:memory-retention}, error in BOSMOS parameter estimates did not converge to zero, showing difficulty in predicting model parameters for PPO and PR models. Improving parameter inference may require modifying priors to encourage more diverse behaviours and selecting more descriptive experimental responses. Finally, BOSMOS outperformed MINEBED across all performance metrics after only one design trial, with the model predictive accuracy showing a large difference, establishing BOSMOS as a clear favourite approach for this task. 

\begin{figure}[h!]
    \centering
    \includegraphics[width=\textwidth]{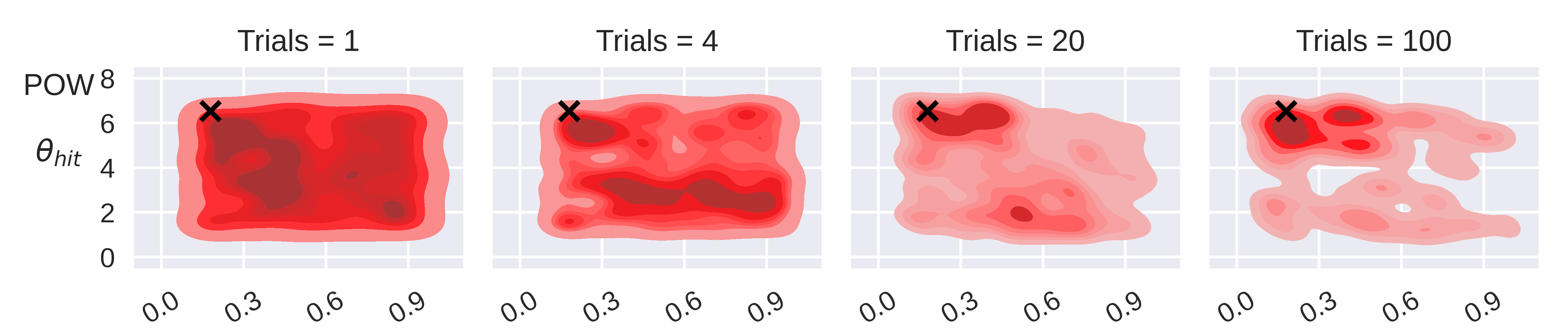}
    \includegraphics[width=\textwidth]{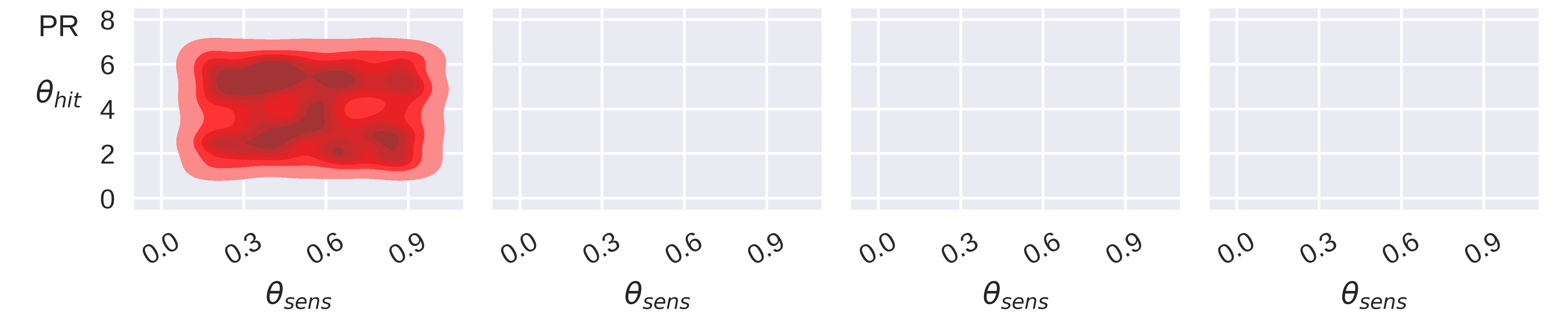}
    \includegraphics[width=\textwidth]{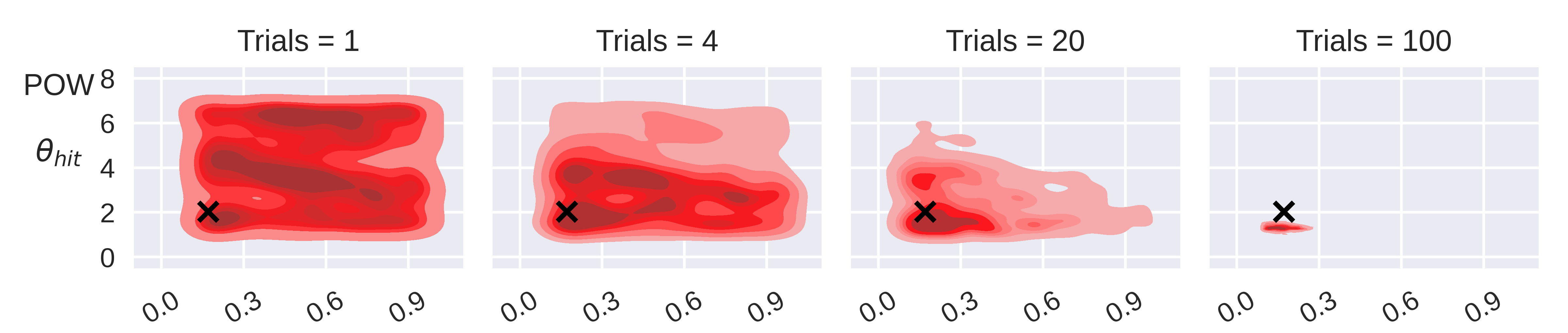}
    \includegraphics[width=\textwidth]{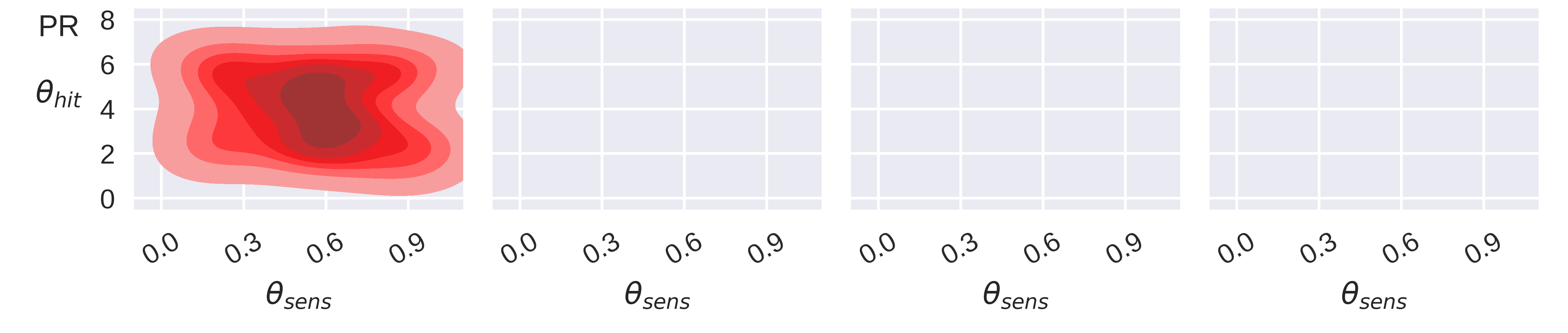}
    \caption{An example of evolution of the posterior approximation in each of the models tested resulting from BOSMOS in the signal detection task. The last bottom row panels are empty as in both cases the posterior probability of the PR model becomes negligible, so that the particle approximation of this posterior does not contain any more particle. The true value of the parameters is indicated by the cross and the true model is POW in both cases. BOSMOS successfully identified the ground-truth model in both cases: all posterior density (shaded area) has concentrated there by $20$ trials, and no more particle exists in the other model. However, only in the first example (top panel) did the ground-truth parameter values (cross) fall inside the $90\%$ confidence interval, indicating some inconsistency in terms of the posterior convergence towards the ground-truth. The axes correspond to the model parameters: sensor-noise (x-axis) and hit value (y-axis); $\theta_\text{low}$ and $\theta_\text{len}$ of the PR model are omitted to simplify visualization.}
    \label{fig:comp-post}
\end{figure}

An example of posterior distributions returned by BOSMOS is demonstrated in Figure \ref{fig:comp-post}. Despite overall positive results, there are occasional cases in a population of synthetic participants, where BOSMOS fails to converge to the ground-truth. The same problem can be observed with MINEBED, as demonstrated in Appendix D. These findings may be attributed to poor identifiability of the signal detection models, suggested earlier in the memory task, but also due to the approximation inaccuracies accumulated over numerous trials. Since both methods operate in a LFI setting, some inconsistency between replicating the target behaviour and converging to the ground-truth parameters is to be expected when the models are poorly identifiable.

\subsection{Risky choice}
\label{sec:risky-choice}

Risky choice problems are typical tasks used in psychology, cognitive science and economics to study attitudes towards uncertainty.
Specifically, risk refers to `quantifiable' uncertainty, where a decision-maker is aware of probabilities associated with different outcomes \citep{knight_risk_1921}.
In risky choice problems, individuals are presented with options that are lotteries (i.e., probability distributions of outcomes). 
For example, a risky choice problem could be a decision between winning 100 euros with a chance of 25\%, or getting 25 euros with a chance of 99\%.
The choice is between two lotteries (100, 0.25; 0, 0.75) and (25, 0.99; 0, 0.01).
The goal of the participant is to maximize the subjective reward of their single choice, so they need to assess the risk associated with outcomes in each lottery.

Several models have been proposed to explain tendencies in these tasks, including normative approaches derived from logic to descriptive approaches based on empirical findings \citep{johnson_2010_dmunderrisk}.
In this paper, we will consider four classic models (following \citealp{cavagnaro2013discriminating}): expected utility theory (EU) \citep{vnm_eu_1990}, weighted expected utility theory (WEU) \citep{chew1983weu}, original prospect theory (OPT) \citep{kahnemantversky1979pt} and cumulative prospect theory (CPT) \citep{tverskykahneman1992cpt}.
The risky choice models we consider consist of a subjective utility objective (characterizing the amount of value an individual attaches to an outcome) and possibly a probability weighting function (reflecting the tendency for non-linear weighting of probabilities). Despite the long history of development, risky choice is still a focus of the ongoing research \citep{begenau2020capital, gachter2022individual, frydman2022efficient}.

The objective is to maximize reward from risky choices. 
Risky choice problems consist of 2 or more options, each of which is described by a set of probability and outcome pairs. 
For each option, the probabilities sum to 1. 
Problems may also have an endowment and/or have multiple stages. 
These variants are not modelled in this version. 
We will use similar implementations as \cite{cavagnaro2013discriminating} to test four models $\mathcal{M}$ with our method: EU, WEU, OPT and CPT.
Each model has its own corresponding parameters $\vtheta_m$.
We consider choice problems where individuals choose between two lotteries $A$ and $B$.
The design space for the risky-choice problems is a combination of designs for lottery $A$ and $B$.
The design space for lottery $A$ is defined as the probabilities of the high and low outcome ($d_{\text{phA}}$ and $d_{\text{plA}}$) in this lottery.
The design space for lottery $B$ is analogous to lottery $A$ ($d_{\text{phB}}$ and $d_{\text{plB}}$).
We assume that there the decisions contain choice stochasticity, which serves as a likelihood for the ADO and LBIRD methods.
The models are implemented as follows.

\paragraph{Choice stochasticity.}
It is typical to assume that individual choices in risky choice problems are not deterministic (i.e., there is choice stochasticity).
We use the following definition for probability of choosing lottery $A$ over $B$ in a choice problem $i$ \citep{cavagnaro2013optimal}:
\begin{numcases}{\phi_i(\textit{A}_i \text{\textbar} \vtheta_m, \epsilon)=}
  \epsilon, & if  $\textit{A}_i \prec \textit{B}_i$\\
  \frac{1}{2}, & if  $\textit{A}_i \sim \textit{B}_i$\\
  1-\epsilon, & if  $\textit{A}_i \succ \textit{B}_i$
\end{numcases}
where $\theta_m$ denotes the model parameters and $\epsilon$ is a value in range [0,0.5] quantifying stochasticity of the choice (with $\epsilon=0$ corresponding to a deterministic choice).
Whether lottery $\textit{A}$ is preferred is determined using the utilities defined for each model separately.

\paragraph{EU.}
Following \cite{cavagnaro2013discriminating}, we specify EU using indifference curves on the Marschak-Machina (MM) probability triangle.
Lottery $\textit{A}$ consists of three outcomes ($x_\text{lA}$, $x_\text{mA}$, $x_\text{hA}$), and associated probabilities ($p_\text{lA}$, $p_\text{mA}$, $p_\text{hA}$).
Lottery $A$ can be represented using a right triangle (MM) with two of the probabilities as the plane ($p_\text{lA}$ and $p_\text{hA}$ as $x$ and $y$ axes, respectively).
Hence, the design space for lottery $A$ consists of only the high and low probability ($d_\text{plA}$ and $d_\text{phA}$).
Lottery $B$ can be represented on the triangle similarly (using $d_\text{plB}$ and $d_\text{phB}$).
Then, indifference curves can be drawn on this triangle, as their slope represents the marginal rate of substitution between the two probabilities.
EU is defined using indifference curves that all have the same slope $\theta_a \in \theta_{\text{EU}}$.
If lottery $B$ is riskier, $A \succ B$, if 
$\mid d_{\text{phB}}-d_{\text{phA}} \mid / \mid d_{\text{plB}}-d_{\text{plA}} \mid < \theta_a$.
We ask to turn to \cite{cavagnaro2013discriminating} for a more comprehensive explanation of this modelling approach.

\paragraph{WEU.}
WEU is also defined using the MM-triangle, as per \cite{cavagnaro2013discriminating}. 
In contrast to EU, the slope of the indifference curves varies across the MM-triangle for WEU.
This is achieved by assuming that all the indifference curves intersect at a point ($\theta_x$, $\theta_y$) outside the MM-triangle, where $[\theta_x$, $\theta_y] \in \theta_{\text{WEU}}$.
Then, $\textit{A}\succ \textit{B}$, if 
$\mid d_{\text{phA}} - \theta_y \mid / \mid d_{\text{plA}} - \theta_x \mid > \mid d_{\text{phB}} - \theta_y \mid / \mid d_\text{plB}- \theta_x \mid $.

\paragraph{OPT.}
In contrast to EU and WEU, OPT assumes that both the outcomes $x$ and probabilities $p$ have specific editing functions $v$ and $w$, respectively.
Assuming that for lottery $\textit{A}$, $v(x_\text{low}^{\textit{A}})=0$ and $v(x_\text{high}^{\textit{A}})=1$, the utility objectives in OPT can be defined using $v(x_\text{middle}^{\textit{A}})$ as a parameter $\theta_v$
\begin{numcases}{u(\textit{A})=}
  w(d_\text{phA})\cdot 1 + \theta_v \cdot(1-w(d_{\text{phA}})), & if $d_\text{plA}=0$\\
  w(d_\text{phA}) \cdot 1 + w(1-d_\text{phA}-d_\text{plA}) \cdot \theta_v, & otherwise.
\end{numcases}

Utility $u(B)$ for lottery \textit{B} can be calculated analogously, and $A_i \succ B_i$ if $u(A) > u(B)$.
The probability weighting function $w(\cdot)$ used is the original work by \cite{tverskykahneman1992cpt} is
\begin{equation}
    w(p)=\frac{p^{\theta_r}}{(p^{\theta_r}+(1-p)^{\theta_r})^{(1/{\theta_r})}},
\end{equation}
where $\theta_r$ is a parameter describing the shape of the function.
Thus, OPT has two parameters $[\theta_v$, $\theta_r] \in \theta_{\text{OPT}}$, describing the subjective utility of the middle outcome and the shape of the probability weighting function, respectively.

\paragraph{CPT.}
CPT is defined similarly to OPT, however, the subjective utilities $u$ for lottery $A$ are calculated using
\begin{equation}
    u(A) = w(d_\text{phA})\cdot 1 +(w(1-d_\text{plA})-w(d_\text{phA}))\cdot \theta_v.
\end{equation}

Utility $u(B)$ for lottery $B$ is calculated similarly and $[\theta_v$, $\theta_r] \in \theta_{\text{CPT}}$.
We use the following priors for the parameters of models
\begin{align}
    \theta_a &\sim \text{Unif}(0, 10), & \theta_v &\sim \text{Unif}(0, 1), \\
    \theta_r &\sim \text{Unif}(0.01, 1), & \theta_x &\sim \text{Unif}(-100, 0), \\
    \theta_y &\sim \text{Unif}(-100, 0), & \theta_\epsilon &\sim \text{Unif}(0, 0.5),
\end{align}
with the design proposal distributions
\begin{align}
    d_\text{plA} &\sim \text{Unif}(0,1), & d_\text{phA} &\sim \text{Unif}(0,1), \\
    d_\text{plB} &\sim \text{Unif}(0,1), & d_\text{phB} &\sim \text{Unif}(0,1).
\end{align}
Please note that $d_\text{pmA}$ and $d_\text{pmB}$ can be calculated analytically from $d_\text{pmA} = 2 - d_\text{plA} - d_\text{phA}$, after which the designs for the same lottery ($d_\text{plA}$, $d_\text{pmA}$, $d_\text{phA}$) are normalized, so they are summed to $1$ (and similar for the lottery B).

\subsubsection{Results}

The risky choice task comprises four computational models, which significantly expand the space of models and makes it much more computationally costly than the memory task. Despite the larger model space, BOSMOS maintains its position as a preferred LFI approach to model selection, most notably when compared to the parameter estimation error of MINEBED from Figure \ref{fig:res-20}. With more models, BOSMOS's performance advantage over MINEBED grows, with BOSMOS exhibiting higher scalability for larger model spaces.

It is crucial to note that having several candidate models reduces model prediction accuracy by the LFI approaches, thus we recommend reducing the number of candidate models as low as feasible. In terms of performance, BOSMOS is comparable to ground-truth likelihood approaches during the first four design trials, as shown in Figure \ref{fig:conv}, since it is significantly easier to minimize uncertainty early in the trials. Similarly to the memory task, the error of LFI approximation becomes more apparent as the number of trials rises, as evidenced by comparing BOSMOS to ADO for the behavioural fitness error and model predictive accuracy. In terms of the parameter estimate error, BOSMOS performs marginally better than ADO. 

Finally, BOSMOS has a relatively low runtime cost, especially compared to other methods (about one minute per design trial). 
This brings adaptive model selection closer to being applicable to real-world experiments in risky choice.
The proposed method can be useful in online experiments that include lag times between trials, for instance, in assessing investment decisions (e.g., \citealp{camerer2004prospect,gneezy1997experiment}) or game-like settings (e.g., \citealp{bauckhage2012players, putkonen2022suitable, viljanen2017playtime}) where the participant waits between events.

\section{Discussion}

In this paper, we proposed a simulator-based experimental design method for model selection, BOSMOS, that does design selection for model and parameter inference at a speed orders of magnitude higher than other methods, bringing the method closer to online design selection. This was made possible with newly proposed approximation of the model likelihood and simulator-based utility objective. Despite needing orders of magnitude fewer simulations, BOSMOS significantly outperformed LFI alternatives in the majority of cases, while being orders of magnitude faster, bringing the method closer to an online inference tool. Crucially, the time between experiment trials was reduced to less than a minute. 
Whereas in some settings this time between trials may be too long, BOSMOS is a viable tool in experiments where the tasks include a lag time, for instance, in studies of language learning (e.g., \citealp{gardner1997towards, nioche2021improving}) and task interleaving (e.g., \citealp{payne2007discretionary, brumby2009interleaving,gebhardt2021hierarchical,katidioti2014happens}). Moreover, our code implementation represents a proof of concept and was not fully optimized for maximal efficiency: in particular, a parallel implementation that exploits multiple cores and batches of simulated experiments would enable additional speedups \citep{wu2016parallel}.
As an interactive and sample-efficient method, BOSMOS can help reduce the number of required experiments. This can be of interest to both the subject and the experimenter. In human trials it allows for faster interventions (e.g. adjusting the treatment plan) in critical settings such as ICUs or RCTs. However, it can also have detrimental applications, such as targeted advertising and collecting personal data, therefore the principles and practices of responsible AI \citep{dignum2019responsible, arrieta2020explainable} also have to be taken into account in applying our methodology.

There are at least two remaining issues left for future work. The first issue we witnessed in our experiments is that the accuracy of behaviour imitation does not necessarily correlate with the convergence to ground-truth models. This usually happens due to poor identifiability in the model-parameter space, which may be quite prevalent in current and future computational cognitive models, since they are all designed to explain the same behaviour. Currently, the only way to address this problem is to use Bayesian approaches, such as BOSMOS, that quantify the uncertainty over the models and their parameters. The second issue is the consistency of the method: in selecting only the most informative designs, the methods may misrepresent the posterior and return an overconfident posterior. This bias may occur, for example, due to a poor choice of priors or summary statistics \citep{nunes2010optimal, fearnhead2012constructing} for the collected data (when the data is high-dimensional). Ultimately, these issues do not hinder the goal of automating experimental designs, but introduce the necessity for a human expert, who would ensure that the uncertainty around estimated models is acceptable, and the design space is sufficiently explored to make final decisions.

Future work for simulator-based model selection in computational cognitive science needs to consider adopting hierarchical models, accounting for the subjects' ability to adapt or change throughout the experiments, and incorporating amortized non-myopic design selection. 
A first step in this direction would be to study hierarchical models \citep{kim2014hierarchical} which would allow adjusting prior knowledge for populations and expanding the theory development capabilities of model selection methods from a single individual to a group level.  
We could also remove the assumption on the stationarity of the model by proposing a dynamic model of subjects' responses which adapts to the history of previous responses and previous designs, which is more reasonable in longer settings of several dozens of trials.
Lastly, amortized non-myopic design selections \citep{blau2022optimizing} would even further reduce the wait time between design proposals, as the model can be pre-trained before experiments, and would also improve design exploration by encouraging long-term planning of the experiments. Addressing these three potential directions may have a synergistic effect on each other, thus expanding the application of simulator-based model selection in cognitive science even further.

\subsubsection*{Supplementary information}

The article has the following accompanying supplementary materials:
\begin{itemize}
    \item Appendix A shows the validity of the approximation of the entropy gain for the design selection rule;
    \item Appendix B details the algorithm for the proposed BOSMOS method;
    \item Appendix C contains tables with full experimental results, which shows additional design evaluation points;
    \item Appendix D showcases a side-by-side comparison of the posterior evolution resulting from BOSMOS and MINEBED for the signal detection task. 
\end{itemize}

\subsubsection*{Acknowledgements}

This work was supported by the Academy of Finland (Flagship programme: Finnish Center for Artificial Intelligence FCAI; grants 328400, 345604, 328400, 320181). AP and SC were  funded by the Academy of Finland projects BAD (Project ID: 318559) and Human Automata (Project ID: 328813). AP was additionally funded by Aalto University School of Electrical Engineering. SC was also funded by Interactive Artificial Intelligence for Research and Development (AIRD) grant by Future Makers. SK was supported by the Engineering and Physical Sciences Research Council (EPSRC; Project ID: EP-W002973-1). Computational resources were provided by the Aalto Science-IT Project.

\subsubsection*{Code Availability}

All code for replicating the experiments is available at \href{https://github.com/AaltoPML/BOSMOS}{https://github.com/AaltoPML/BOSMOS}.

\bibliographystyle{plainnat}
\bibliography{ref}

\newpage
\appendix

\section{Approximation of the entropy gain for design selection}

Since the expected values in the entropy gain  from Equation \eqref{eq:designloss} are not tractable, we rely on a Monte-Carlo estimation of these quantities. We focus on the first term, as the second one only features one of the approximations.

Following the SMC framework \citep{del2006sequential}, we propose to sequentially update a particle population between trials. This allows the online update of the posterior along the experiment.

At each step we estimate the distribution $p(m,\vtheta_m \mid \data_{1:t})$ with a population of $N_1$ particles $ (m^i,\vtheta_m^i)$. Following importance sampling, we know that $\sum_{i=1}^{N_1} w_i\delta_{(m^i,\vtheta_m^i)}$ converges to the distribution associated with density $p(m,\vtheta_m \mid \data_t)$. This population of particles then evolves according to the SMC algorithm \citep{del2006sequential}.

To estimate the expected value in Equation \eqref{eq:designloss}, we use a standard Monte-Carlo estimation, making use of the particles at time $t$ as an estimation of the posterior at time $t$. For each particle, we simulate $N_2$ vectors $\vx_t'^i(j) \sim p( \cdot \mid \vd_t,\vtheta_m^i,m^i)$. The convergence to prove is then:

\begin{equation}
\sum_{i,j} (N_2)^{-1}w_i \hat{H}(\vx_t'^i(j) \mid m^i,\vtheta_m^i) \rightarrow_{N_1,N_2 \to \infty} \mathbb{E}_{q(m, \vtheta_m \mid \data_{t-1})} [H(\vx_t' \mid m, \vtheta_m)], \label{eq:entropy_approx}
\end{equation} 
where $\hat{H}$ is a modified version of the entropy. Note that it is also possible to estimate the gradient of this quantity with respect to the design. If we truly computed $H(\vx'_t)$ as the entropy with respect to the measure $\sum_i \delta_{x'_i}$, it would lead to constantly null value. Thus, we decide to use a kernel approximation of the distribution: $\hat{H}(\vx_t'^i(j) \mid m^i,\vtheta_m^i)= H(\sum_i \mathcal{N}(\cdot \mid x'_i(j), \sigma_{N_2}) \mid m^i,\vtheta_m^i)$, with $\sigma_{N_2} \rightarrow_{N_2 \to \infty} 0$.

The convergence of the estimator in Equation \ref{eq:entropy_approx} to the true entropy requires two results.

First, the convergence in $N_2$:
\[ \sum_{i=1}^{N_1}\sum_{j=1}^{N_2} (N_2)^{-1}w_i \hat{H}(\vx_t'^i(j) \mid m^i,\vtheta_m^i) \rightarrow_{N_2 \to \infty } \sum_{i=1}^{N_1}w_i \hat{H}(p( \cdot \mid \vd_t,\vtheta_m^i,m^i) \mid m^i,\vtheta_m^i), \]
using the convergence of $\mathcal{N}(\cdot \mid x'_i(j), \sigma_{N_2})$ to $\delta_{x'_i(j)}$ in distribution and the law of large numbers. Second, the convergence for $N_1 \to \infty$ comes from the results on SMC \citep{del2006sequential}.

\section{Algorithms}

\begin{algorithm}[H]
   \caption{Bayesian optimization for simulator-based model selection}
   \label{alg:lfi-param}
\begin{algorithmic}
    \STATE {\bfseries Input:} prior over models $p(m)$ and parameters ${p(\vtheta_m \mid m)}_m$; set of all models $\mathcal{M} = \{ p(\vx \mid \vtheta_m, m, d) \}$; design budget $N_d$; total number of particles $N_q$;
    \STATE {\bfseries Output:} selected model $m'$ and its parameters $\vtheta_m'$;
    \STATE
    \STATE initialize current beliefs from the priors: 
    \STATE $\quad q(m, \vtheta_m) := \{ (m', \vtheta_m'): \vtheta_m' \sim p(\vtheta_m \mid m'), m' \sim p(m)\}_{i=0}^{N_q}$;
    \STATE initialize an empty set for the collected data: $\data_0 = \{\}$;
    \FOR{ $i := 1:N_d$ }
        \STATE get the design $\vd'$ with Equation~\eqref{eq:heurloss_prat};
        \STATE collect the data $\vx'$ at the design location $\vd'$ and store it in $\data_{i}$;
        \FOR{$m' \in \mathcal{M}$}
            \STATE get the likelihood $\mathcal{L}_{\epsilon_{m'}}(\vx_i \mid \vtheta_{m'})$ with Equation~\eqref{eq:liklhd};
        \ENDFOR
        \STATE get the marginal likelihood $\mathcal{L}(\vx_i \mid m, \data_{i-1})$ with Equation~\eqref{eq:kernel-approx};
        \STATE update $q(\vtheta_m, m \mid \data_{i})$ with Equation~\eqref{eq:update};
    \ENDFOR
    
    \STATE apply the decision rule (e.g. MAP): 
    \STATE $\quad m' = \text{arg max}_m \sum_{\vtheta_m} q(\vtheta_m, m \mid \data_{N_d})$;  
    \STATE $\quad \vtheta_m' = \text{arg max}_{\vtheta_m} q(\vtheta_m \mid m, \data_{N_d})$;
\end{algorithmic}
\end{algorithm}

\section{Full experimental results}
\label{sec:analysis}

We provide full experimental results data in Tables \ref{tab:behconv}-\ref{tab:modelaccuracy}, where evaluations of performance metrics were made after different numbers of design iterations. Moreover, we report time costs of running $100$ design trials for each method in Table \ref{tab:timecost}, where our method was 80-100 times faster than the other LFI method, MINEBED. The rest of the section discusses three additional minor points with relation to the performance of ADO for the demonstrative example, the bias in the model space for the memory task, and results of testing two decision rules in the risky choice. 

As we have seen in the main text, MINEBED had the fastest behavioural fitness convergence rate in the demonstrative example, while BOSMOS was the close second. Hence, ADO had the slowest convergence rate among design optimization methods for behavioural fitness (Table \ref{tab:behconv}) and also for parameter estimation (Table \ref{tab:parconv}). This result is somewhat counterintuitive, as we expected ADO, with its ground-truth likelihood and design optimization, to be the fastest to converge. Since the only other factor influencing this outcome, Bayesian updates, had access to the ground-truth and avoided LFI approximations, suboptimal designs are likely to blame for the poor performance. This problem is likely to be mitigated by expanding the size of the grid used by ADO to calculate the utility objective. However, expanding it would likely increase ADO's convergence at the expense of more calculations; therefore we aimed to get its running time closer to that of BOSMOS, so both methods could be fairly compared.

In the results of model selection for the memory retention task discussed in the main text, MINEBED showed a marginally better average model accuracy than BOSMOS. However, upon closer inspection (Table \ref{tab:modelaccuracy}), this accuracy can be solely attributed to the strong bias towards the POW model; the other approaches show it as well, albeit less dramatically. This suggests that the two models in the memory task are separable, but the EXP model cannot likely replicate parts of the behaviour space that the POW model can, resulting in this skewness towards the more flexible model. This is also more broadly related to non-identifiability: since these cognitive models were designed to explain the same target behaviour, it is inevitable that there will be an overlap in their response (or behavioural data) space, complicating model selection.

Since the risky choice model had four models of varied complexity, we experimented with two distinct decision-making rules for estimating the models and parameters: the default MAP and Bayesian information criterion (BIC) \citep{schwarz1978estimating, vrieze2012model}. Both decision-making rules include a penalty for the size of the model (artificially for BIC, and by definition for MAP). Interestingly, the results are the same for both decision-making rule, indicating that the EU model cannot be completely replaced by a more flexible model. BIC's slightly superior parameter estimates for the BOSMOS technique is most likely explained by the poorer model prediction accuracy. Nevertheless, the BIC rule remains a viable option for model selection in situations when there is a risk of having a more flexible and complex model alongside few-parameter alternatives, despite being less supported theoretically.

\begin{table}
    \centering
    \begin{tabular}{cccccc}
         \hline
         \textbf{Methods} & \multicolumn{5}{c}{\textbf{Tasks: number of design trials}} \\ \hline
         &  \multicolumn{5}{c}{\textbf{Demonstrative example}} \\
         & 1 trial & 2 trials & 4 trials & 20 trials & 100 trials \\ 
         ADO & 0.03 $\pm$ 0.03 & 0.02 $\pm$ 0.02 & 0.02 $\pm$ 0.02 & 0.01 $\pm$ 0.01 & -  \\
         MINEBED &  0.02 $\pm$ 0.07 &  0.01 $\pm$ 0.00  & 0.01 $\pm$ 0.00 & 0.01 $\pm$ 0.00 & -   \\
         BOSMOS & 0.05 $\pm$ 0.07 & 0.01 $\pm$ 0.01 & 0.01 $\pm$ 0.00 & 0.01 $\pm$ 0.00 & - \\
         LBIRD &  0.36 $\pm$ 0.24 & 0.33 $\pm$ 0.25 & 0.29 $\pm$ 0.24 & 0.14 $\pm$ 0.18  & -  \\
         Prior & \multicolumn{5}{c}{Baseline for 0 trials: 0.33 $\pm$ 0.3 } \\
         
         &  \multicolumn{5}{c}{\textbf{Memory retention}} \\
         & 1 trial & 2 trials & 4 trials & 20 trials & 100 trials \\ 
         ADO & 0.20 $\pm$ 0.16 & 0.17 $\pm$ 0.14 & 0.15 $\pm$ 0.10 & 0.07 $\pm$ 0.06 & 0.05 $\pm$ 0.03  \\
         MINEBED & 0.27 $\pm$ 0.22 & 0.24 $\pm$ 0.19 &  0.24 $\pm$ 0.19 &  0.23 $\pm$ 0.15  & 0.23 $\pm$ 0.17 \\
         BOSMOS & 0.24 $\pm$ 0.19 & 0.19 $\pm$ 0.16 & 0.17 $\pm$ 0.14 & 0.15 $\pm$ 0.13 & 0.13 $\pm$ 0.11 \\
         LBIRD & 0.20 $\pm$ 0.17 & 0.17 $\pm$ 0.15 & 0.14 $\pm$ 0.11 & 0.08 $\pm$ 0.06 & 0.05 $\pm$ 0.03 \\
         Prior & \multicolumn{5}{c}{Baseline for 0 trials: 0.33 $\pm$ 0.47 } \\
         
         &  \multicolumn{5}{c}{\textbf{Signal detection}} \\
         & 1 trial & 2 trials & 4 trials & 20 trials & 100 trials \\ 
         MINEBED & 0.27 $\pm$ 0.24 & 0.24 $\pm$ 0.20 & 0.23 $\pm$ 0.17 & 0.21 $\pm$ 0.18 & 0.20 $\pm$ 0.17 \\
         BOSMOS & 0.25 $\pm$ 0.21 & 0.20 $\pm$ 0.17 & 0.17 $\pm$ 0.15 & 0.17 $\pm$ 0.15 & 0.15 $\pm$ 0.12 \\
         Prior & \multicolumn{5}{c}{Baseline for 0 trials: 0.40 $\pm$ 0.49 } \\
         
         &  \multicolumn{5}{c}{\textbf{Risky choice}} \\
         & 1 trial & 2 trials & 4 trials & 20 trials & 100 trials \\ 
         ADO & 0.32 $\pm$ 0.11 & 0.30 $\pm$ 0.13 & 0.27 $\pm$ 0.12 & 0.14 $\pm$ 0.08 & 0.07 $\pm$ 0.04 \\
         MINEBED & 0.30 $\pm$ 0.11 & 0.31 $\pm$ 0.12 & 0.26 $\pm$ 0.12 & 0.21 $\pm$ 0.12 & 0.22 $\pm$ 0.13  \\
         BOSMOS & 0.26 $\pm$ 0.11 & 0.23 $\pm$ 0.12 & 0.24 $\pm$ 0.13 & 0.18 $\pm$ 0.11 & 0.14 $\pm$ 0.08 \\
         MINEBED (BIC) & 0.25 $\pm$ 0.11 & 0.26 $\pm$ 0.13 & 0.23 $\pm$ 0.11 & 0.21 $\pm$ 0.12 & 0.22 $\pm$ 0.13  \\
         BOSMOS (BIC) & 0.24 $\pm$ 0.11 & 0.24 $\pm$ 0.13 & 0.23 $\pm$ 0.11 & 0.19 $\pm$ 0.12 & 0.15 $\pm$ 0.10 \\
         LBIRD & 0.31 $\pm$ 0.12  & 0.30 $\pm$ 0.13 & 0.26 $\pm$ 0.13  & 0.14 $\pm$ 0.07 & 0.08 $\pm$ 0.03  \\
         Prior & \multicolumn{5}{c}{Baseline for 0 trials: 0.44 $\pm$ 0.50 } \\
    \end{tabular}
    \caption{Convergence of behavioural fitness error $\eta_\text{b}$ (mean $\pm$ std. across $100$ simulated participants) for comparison methods (rows) with increased number of trials (columns).}
    \label{tab:behconv}
\end{table}

\begin{table}
    \centering
    \begin{tabular}{cccccc}
         \hline
         \textbf{Methods} & \multicolumn{5}{c}{\textbf{Tasks: number of design trials}} \\ \hline
         
         &  \multicolumn{5}{c}{\textbf{Demonstrative example}} \\
         & 1 trial & 2 trials & 4 trials & 20 trials & 100 trials \\ 
         ADO & 0.05 $\pm$ 0.06 & 0.04 $\pm$ 0.05 & 0.03 $\pm$ 0.04  & 0.01 $\pm$ 0.01  & -  \\
         MINEBED & 0.00 $\pm$ 0.02 & 0.00 $\pm$ 0.00 & 0.00 $\pm$ 0.00 & 0.00 $\pm$ 0.00  & -  \\
         BOSMOS & 0.05 $\pm$ 0.07 & 0.01 $\pm$ 0.02 & 0.00 $\pm$ 0.00 & 0.00 $\pm$ 0.00 & - \\
         LBIRD & 0.34 $\pm$ 0.22 & 0.32 $\pm$ 0.24 & 0.29 $\pm$ 0.25 & 0.17 $\pm$ 0.18  & -  \\
         Prior & \multicolumn{5}{c}{Baseline for 0 trials: 0.33 $\pm$ 0.23 } \\
         
         &  \multicolumn{5}{c}{\textbf{Memory retention}} \\
         & 1 trial & 2 trials & 4 trials & 20 trials & 100 trials \\ 
         ADO & 0.25 $\pm$ 0.21 & 0.25 $\pm$ 0.20 & 0.23 $\pm$ 0.19 & 0.19 $\pm$ 0.14  & 0.14 $\pm$ 0.12  \\
         MINEBED & 0.47 $\pm$ 0.38 & 0.46 $\pm$ 0.38 & 0.46 $\pm$ 0.39  & 0.48 $\pm$ 0.40  & 0.48 $\pm$ 0.43  \\
         BOSMOS & 0.29 $\pm$ 0.21 & 0.27 $\pm$ 0.22 & 0.28 $\pm$ 0.20 & 0.27 $\pm$ 0.22 & 0.29 $\pm$ 0.22 \\
         LBIRD & 0.25 $\pm$ 0.21 & 0.25 $\pm$ 0.20 & 0.22 $\pm$ 0.18 & 0.22 $\pm$ 0.20  & 0.15 $\pm$ 0.13  \\
         Prior & \multicolumn{5}{c}{Baseline for 0 trials: 0.33 $\pm$ 0.20 } \\
         
         &  \multicolumn{5}{c}{\textbf{Signal detection}} \\
         & 1 trial & 2 trials & 4 trials & 20 trials & 100 trials \\
         MINEBED & 0.60 $\pm$ 0.24 & 0.56 $\pm$ 0.28 & 0.43 $\pm$ 0.34 & 0.45 $\pm$ 0.35 & 0.49 $\pm$ 0.34  \\ 
         BOSMOS & 0.37 $\pm$ 0.22 & 0.35 $\pm$ 0.19 & 0.36 $\pm$ 0.21 & 0.35 $\pm$ 0.20 & 0.35 $\pm$ 0.19 \\
         Prior & \multicolumn{5}{c}{Baseline for 0 trials: 0.35 $\pm$ 0.20 } \\
         
         &  \multicolumn{5}{c}{\textbf{Risky choice}} \\
         & 1 trial & 2 trials & 4 trials & 20 trials & 100 trials \\ 
         ADO & 0.42 $\pm$ 0.22  & 0.44 $\pm$ 0.23 & 0.43 $\pm$ 0.22 & 0.33 $\pm$ 0.24  & 0.26 $\pm$ 0.23 \\
         MINEBED & 0.86 $\pm$ 0.28 & 0.87 $\pm$ 0.27 & 0.81 $\pm$ 0.32 & 0.76 $\pm$ 0.36  & 0.86 $\pm$ 0.29  \\
         BOSMOS & 0.41 $\pm$ 0.25 & 0.40 $\pm$ 0.21 & 0.45 $\pm$ 0.26 & 0.29 $\pm$ 0.25 & 0.21 $\pm$ 0.23 \\
         MINEBED (BIC) & 0.85 $\pm$ 0.33 & 0.83 $\pm$ 0.31 & 0.87 $\pm$ 0.30 & 0.86 $\pm$ 0.31 & 0.84 $\pm$ 0.35 \\
         BOSMOS (BIC) & 0.27 $\pm$ 0.23 & 0.26 $\pm$ 0.19 & 0.40 $\pm$ 0.26 & 0.23 $\pm$ 0.22 & 0.21 $\pm$ 0.22 \\
         LBIRD & 0.51 $\pm$ 0.25 & 0.48 $\pm$ 0.24 & 0.40 $\pm$ 0.24 & 0.35 $\pm$ 0.22  & 0.24 $\pm$ 0.19  \\
         Prior & \multicolumn{5}{c}{Baseline for 0 trials: 0.48 $\pm$ 0.25 } \\
    \end{tabular}
    \caption{Convergence of parameter estimation error $\eta_\text{p}$ (mean $\pm$ std.  across $100$ simulated participants) when the model is predicted correctly for comparison methods (rows) with increased number of trials (columns).}
    \label{tab:parconv}
\end{table}

\begin{table}
    \centering
    \begin{tabular}{cccccc}
         \hline
         \textbf{Methods} & \multicolumn{5}{c}{\textbf{Tasks: number of design trials}} \\ \hline
         
         & \multicolumn{5}{c}{\textbf{Demonstrative example (PM, NM)}} \\
         & 1 trial & 2 trials & 4 trials & 20 trials & 100 trials \\ 
         ADO & 0.98, 0.96 & 0.98, 1.00 & 0.98, 1.00 & 1.00, 1.00 & -  \\
         MINEBED & 0.93,  0.98 & 0.98, 1.00 & 0.98, 1.00 & 0.98, 1.00 & -  \\
         BOSMOS & 0.87, 0.83 & 1.00, 0.98 & 1.00, 0.98 & 1.00, 1.00 & - \\
         LBIRD & 0.51, 0.48 & 0.56, 0.59 & 0.60, 0.65 & 0.82, 0.87 & -  \\
         Prior & \multicolumn{5}{c}{Baseline for 0 trials: 0.5, 0.5} \\
         
         & \multicolumn{5}{c}{\textbf{Memory retention (POW, EXP)}} \\
         & 1 trial & 2 trials & 4 trials & 20 trials & 100 trials \\ 
         ADO & 0.43, 0.89 & 0.61, 0.80 & 0.74, 0.76 & 0.91, 0.78 & 0.96, 0.82 \\
         MINEBED & 0.53, 0.68 & 0.60, 0.55 & 0.58, 0.50 & 0.96, 0.11 & 0.93, 0.03  \\
         BOSMOS & 0.30, 0.96 & 0.33, 0.96 & 0.22, 0.96 & 0.26, 0.96 & 0.24, 0.96 \\
         LBIRD & 0.37, 0.91 & 0.48, 0.84 & 0.69, 0.87 & 0.94, 0.70 & 0.98, 0.82  \\
         Prior & \multicolumn{5}{c}{Baseline for 0 trials:  0.5, 0.5} \\
         
         & \multicolumn{5}{c}{\textbf{Signal detection (PPO, PR)}} \\
         & 1 trial & 2 trials & 4 trials & 20 trials & 100 trials \\
         MINEBED & 0.02, 0.54  & 0.21, 0.39 & 0.31, 0.22  & 0.33, 0.17 & 0.33, 0.17 \\
         BOSMOS & 0.77, 0.76 & 0.92, 0.52 & 0.94, 0.46 & 0.92, 0.43 & 0.92, 0.43 \\
         Prior & \multicolumn{5}{c}{Baseline for 0 trials: 0.5, 0.5 } \\
         
         & \multicolumn{5}{c}{\textbf{Risky choice (EU, WEU; OPT, CPT)}} \\
         & 1 trial & 2 trials & 4 trials & 20 trials & 100 trials \\ 
         \multirow{2}{*}{ADO} &  0.26, 0.39; & 0.11, 0.39; & 0.22, 0.64; & 0.07, 0.71; & 0.11, 0.86;  \\
         & 0.13, 0.19 & 0.35, 0.24 & 0.35, 0.38 & 0.52, 0.57 & 0.70, 0.81  \\
         \multirow{2}{*}{MINEBED} & 0.38, 0.19; & 0.50, 0.27; & 0.58, 0.23; & 0.38, 0.38; & 0.38, 0.15;   \\
         & 0.12, 0.29 & 0.29, 0.33 & 0.35, 0.24  & 0.12, 0.05 & 0.18, 0.14  \\
         \multirow{2}{*}{BOSMOS} & 0.48, 0.36; &  0.48, 0.36; & 0.52, 0.50; & 0.56, 0.36; & 0.48, 0.32; \\
         & 0.13, 0.05 & 0.09, 0.14 & 0.13, 0.29 & 0.09, 0.33 & 0.09, 0.33 \\
         \multirow{2}{*}{MINEBED (BIC)} & 0.55, 0.00; & 0.00, 0.14; & 0.55, 0.00; & 0.55, 0.00; & 0.55, 0.00;   \\
         & 0.00, 0.00 & 0.23, 0.12 & 0.00, 0.00 & 0.00, 0.00 & 0.00, 0.00  \\
         \multirow{2}{*}{BOSMOS (BIC)} & 1.00, 0.00; & 0.33, 0.36; & 0.52, 0.43; & 0.59, 0.29; & 0.63, 0.29; \\
         & 0.00, 0.00 & 0.04, 0.00 & 0.00, 0.05 & 0.22, 0.05 & 0.22, 0.10 \\
         \multirow{2}{*}{LBIRD} & 0.33, 0.43; & 0.30, 0.43; & 0.33, 0.43; & 0.33, 0.39; & 0.19, 0.50;  \\
         & 0.17, 0.14 & 0.13, 0.29 & 0.13, 0.38 & 0.39, 0.29 & 0.57, 0.67   \\
         Prior & \multicolumn{5}{c}{Baseline for 0 trials: 0.25, 0.25, 0.25, 0.25 } \\
    \end{tabular}
    \caption{Convergence of model prediction accuracy $\eta_\text{m}$ (proportion of correct predictions of models across $100$ simulated participants) for comparison methods (rows) with increased number of trials (columns).}
    \label{tab:modelaccuracy}
\end{table}

\begin{table}[]
    \centering
    \begin{tabular}{ccccc}
        \hline
         & {\small \textbf{Demonstrative example}} & {\small \textbf{Memory retention}} & {\small \textbf{Signal detection}} & {\small \textbf{Risky choice}}  \\ \hline
         ADO & 10.46 $\pm$ 1.06 & 75.44 $\pm$ 9.04 & - & 134.00 $\pm$ 17.08 \\
         MINEBED & 786.47 $\pm$ 87.63 & 3614.11 $\pm$ 272.79 & 6757.20 $\pm$ 399.90 &  6698.34 $\pm$ 310.16 \\ 
         BOSMOS & 7.65 $\pm$ 0.39 & 35.56 $\pm$ 3.54 & 73.41 $\pm$ 8.94 & 88.32 $\pm$ 5.75 \\
         LBIRD & 3.16 $\pm$ 0.37 & 20.07 $\pm$ 1.32 & - & 35.95 $\pm$ 12.58 \\
    \end{tabular}
    \caption{Empirical time cost (mean $\pm$ std. in minutes across $100$ simulated participants) of applying comparison methods (rows) in cognitive tasks (columns) with 100 sequential designs. ADO and LBIRD for the signal detection task need available likelihoods and therefore cannot be used for this task.}
    \label{tab:timecost}
\end{table}

\section{Posterior evolution examples for BOSMOS and MINEBED}

\begin{figure}
    \centering
    \includegraphics[width=\textwidth]{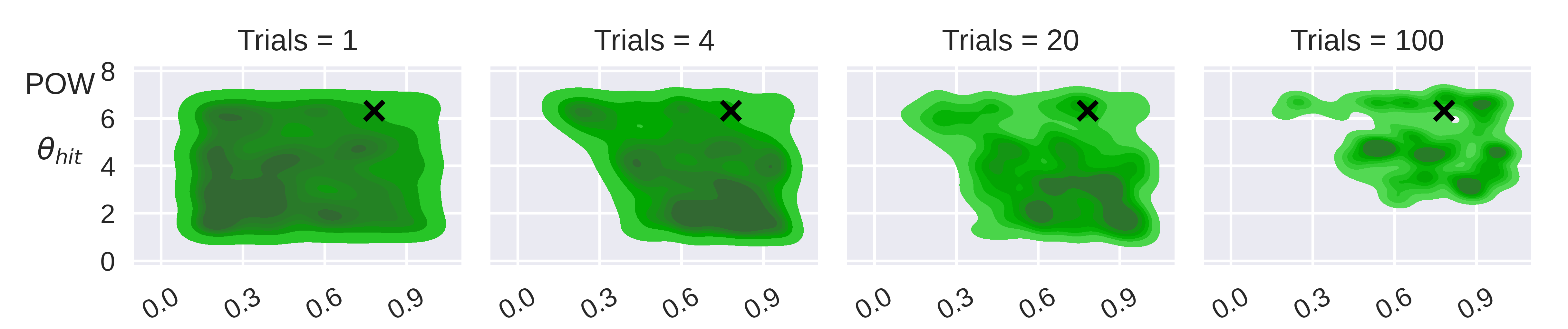}
    \includegraphics[width=\textwidth]{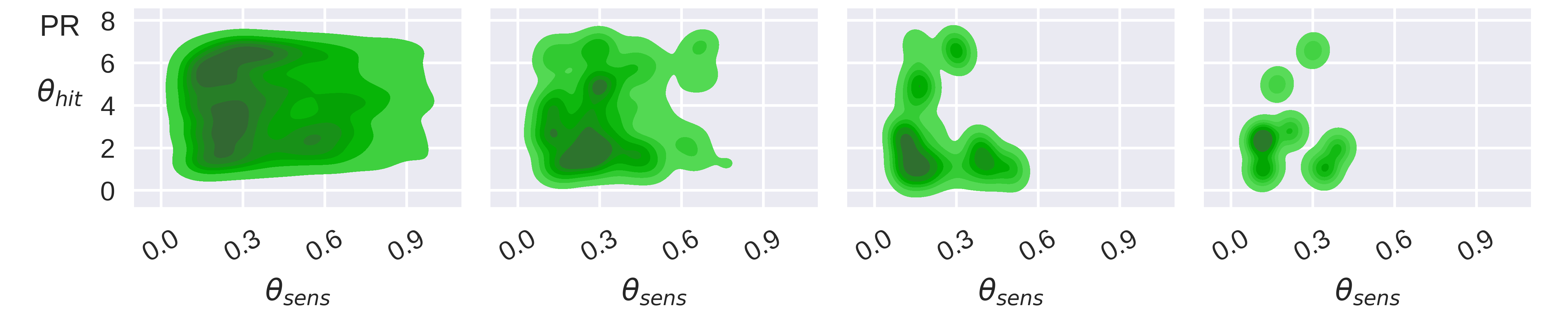}
    \includegraphics[width=\textwidth]{Fig4a.png}
    \includegraphics[width=\textwidth]{Fig4b.png}
    \caption{The first example of evolution of the posterior distribution resulting from MINEBED (green; rows 1 and 2) and BOSMOS (red; rows 3 and 4) for the signal detection task. For each method, the first row shows the distributions of parameters of the POW model (ground-truth), followed by the PR model parameter distributions in the second row. The axes correspond to the model parameters: sensor-noise (x-axis) and hit value (y-axis); $\theta_\text{low}$ and $\theta_\text{len}$ of the PR model are omitted to simplify visualization. The last bottom row panels are empty as in both cases the posterior probability of the PR model becomes negligible, so that the particle approximation of this posterior does not contain any more particle.}
    \label{fig:sm-comp-post-1}
\end{figure}

\begin{figure}
    \centering
    \includegraphics[width=\textwidth]{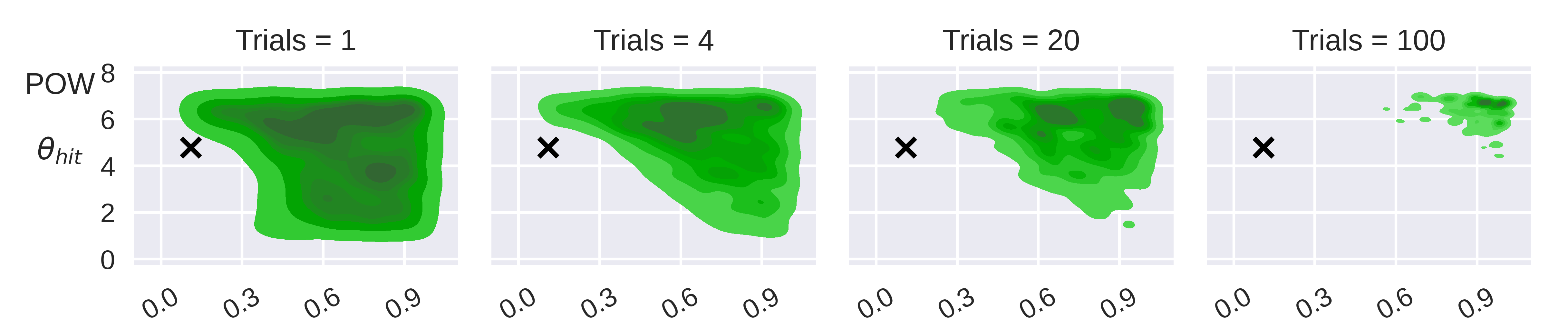}
    \includegraphics[width=\textwidth]{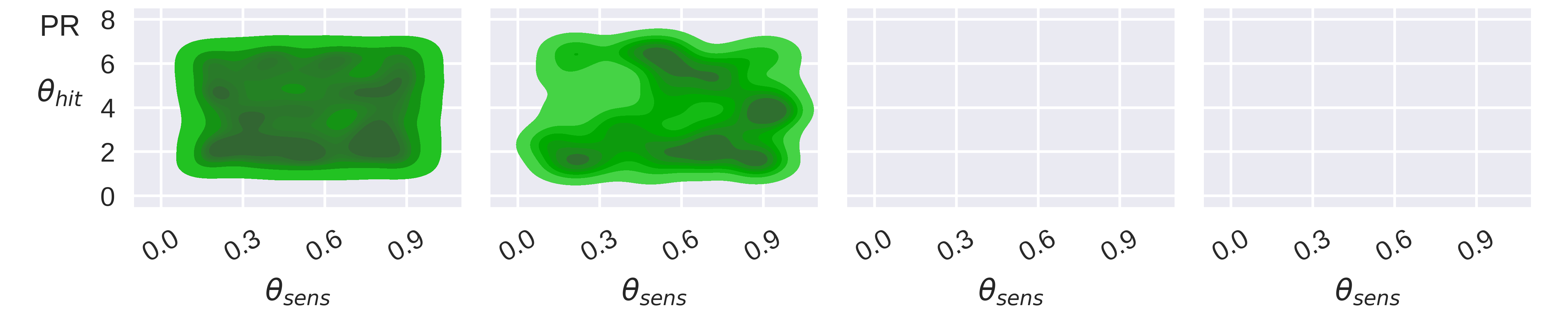}
    \includegraphics[width=\textwidth]{Fig4c.png}
    \includegraphics[width=\textwidth]{Fig4d.png}
    \caption{The second example of evolution of the posterior distribution resulting from MINEBED (green; rows 1 and 2) and BOSMOS (red; rows 3 and 4) for the signal detection task. For each method, the first row shows the distributions of parameters of the POW model (ground-truth), followed by the PR model parameter distributions in the second row. The axes correspond to the model parameters: sensor-noise (x-axis) and hit value (y-axis); $\theta_\text{low}$ and $\theta_\text{len}$ of the PR model are omitted to simplify visualization. The last bottom row panels are empty as in both cases the posterior probability of the PR model becomes negligible, so that the particle approximation of this posterior does not contain any more particle.}
    \label{fig:sm-comp-post-2}
\end{figure}

In Figures \ref{fig:sm-comp-post-1} and \ref{fig:sm-comp-post-2}, we compare posterior distributions returned by MINEBED and BOSMOS for two synthetic participants in the signal detection task. In both examples, the methods have successfully identified the ground-truth POW model, as the majority of the posterior density (shaded area in the figure) has moved to the correct model. Nevertheless, BOSMOS and MINEBED have quite different posteriors, which emphasizes the influence of the design selection strategy on the resulting convergence, as one of the method performs better than the other in each of the provided examples. 

\end{document}